\pdfoutput=1

\documentclass[11pt]{article}

\usepackage{acl}

\usepackage{times}
\usepackage{latexsym}
\usepackage{inconsolata}
\usepackage{graphicx}
\usepackage{subcaption}
\usepackage{booktabs}
\usepackage{multirow}
\usepackage{multicol}
\usepackage{enumitem}
\usepackage{tipa}

\usepackage[T1]{fontenc}

\usepackage[utf8]{inputenc}
\usepackage{amsmath}
\usepackage{amssymb}
\usepackage{linguex}

\usepackage{microtype}

\title{Causal Analysis of Syntactic Agreement Neurons\\in Multilingual Language Models}

\author{Aaron Mueller\textsuperscript{\textnormal{\textipa{S}}}, Yu Xia\textsuperscript{\textnormal{\textipa{C}}}, Tal Linzen\textsuperscript{\textnormal{\textipa{C}}} \\
  \textsuperscript{\textipa{S}}Johns Hopkins University\ \ \ \ \textsuperscript{\textipa{C}}New York University\\
  \texttt{amueller@jhu.edu, yx1675@nyu.edu, linzen@nyu.edu} \\}

\begin{document}
\maketitle
\begin{abstract}
Structural probing work has found evidence for latent syntactic information in pre-trained language models. However, much of this analysis has focused on monolingual models, and analyses of multilingual models have employed correlational methods that are confounded by the choice of probing tasks. In this study, we causally probe multilingual language models (XGLM and multilingual BERT) as well as monolingual BERT-based models across various languages; we do this by performing counterfactual perturbations on neuron activations and observing the effect on models' subject-verb agreement probabilities. We observe where in the model and to what extent syntactic agreement is encoded in each language. We find significant neuron overlap across languages in autoregressive multilingual language models, but not masked language models. We also find two distinct layer-wise effect patterns and two distinct sets of neurons used for syntactic agreement, depending on whether the subject and verb are separated by other tokens. Finally, we find that behavioral analyses of language models are likely underestimating how sensitive masked language models are to syntactic information.
\end{abstract}

\setlength{\Exlabelwidth}{0.25em}
\setlength{\SubExleftmargin}{1.35em}
\setlength{\Extopsep}{0.5\baselineskip}

\section{Introduction}
Syntactic information is necessary for robust generalization in natural language processing tasks (for a case study using the natural language inference task, see \citealt{mccoy2019hans}). The success of pre-trained language models (LMs) such as RoBERTa \citep{roberta} and GPT-3 \citep{gpt3} in many NLP tasks has prompted hypotheses that they accomplish their performance through structural representations induced during pre-training, rather than only lexical or positional representations \cite{manning2020emergent}; behavioral evidence for LMs' syntactic abilities has been found in masked LMs 
(MLMs; \citealp{warstadt2020learning,warstadt2020linguistic,bertgoldberg19}) and autoregressive LMs (ALMs; \citealp{hu2020systematic}). Evidence for structural representations has been reported for multilingual pre-trained LMs \citep{bertgoldberg19,mueller2020cross} and in sequence-to-sequence models \citep{mueller2022blankslate}.

Despite efforts to understand the structural information encoded by pre-trained LMs (\citealp{hewitt2019structural,chi2020mbert_probe,elazar2021amnesic,ravfogel2021counterfactual,finlayson2021causal}; \emph{inter alia}), it remains unclear how and where multilingual models encode this information. Most multilingual probing studies are \emph{correlational} and use dependency parsing or labeling as a proxy task indicative of syntactic information \citep{chi2020mbert_probe,stanczak2022neurons}. This is problematic: Models do not need structural or word order information to achieve high performance on dependency labeling \citep{sinha2021masked}, and training a parametric probing classifier introduces many confounds \citep{hewitt2019selectivity,antverg2022pitfalls}.

Causal probing, however, enables non-parametric analyses of models through counterfactual interventions on inputs or model representations. Causal probing studies have argued for the existence of specific syntactic agreement neurons and units in neural language models \citep{finlayson2021causal,lakretz2019emergence,cao2021diffmask}, but these studies have focused on monolingual models---usually (though not always) in English. Causal methods allow us to make stronger arguments about where and how syntactic agreement is performed in pre-trained LMs, and we can apply them to answer questions about the language specificity and construction specificity of syntactic agreement neurons.

In this study, we extend causal mediation analysis \citep{pearl2001direct,robins2003causaldag,vig2020causal} to multilingual language models, including an autoregressive LM and a masked LM. We also analyze a series of monolingual MLMs across languages. We employ the syntactic interventions approach of \citet{finlayson2021causal} on stimuli in languages typologically related to English, such that we can observe whether there exist syntax neurons that are shared across a set of languages that are all relatively high-resource and grammatically similar.  Our contributions include the following:

\begin{enumerate}[noitemsep]
    \item We causally probe for syntactic agreement neurons in an autoregressive language model, XGLM \citep{xglm}; a masked language model, multilingual BERT \citep{bert}; and a series of monolingual BERT-based models. We find two distinct layer-wise effect patterns, depending on whether the subject and verb are separated by other tokens.
    \item We quantify the degree of neuron overlap across languages and syntactic structures, finding that many neurons are shared across structures and fewer are shared across languages.
    \item We analyze the sparsity of syntactic agreement representations for individual structures and languages, and find that syntax neurons are more sparse in MLMs than ALMs, but also that the degree of sparsity is similar across models and structures.
\end{enumerate}

Our data and code are publicly available.\footnote{\url{https://github.com/aaronmueller/multilingual-lm-intervention}}

\section{Related Work}
\paragraph{Multilingual language modeling.} Multilingual language models enable increased parameter efficiency per language, as well as cross-lingual transfer to lower-resource language varieties \citep{wu19mberteffective}. This makes both training and deployment more efficient when support for many languages is required. A common approach for training multilingual LMs is to concatenate training corpora for many languages into one corpus, often without language IDs \citep{xlmr,bert}.

These models present interesting opportunities for syntactic analysis: Do multilingual models maintain similar syntactic abilities despite a decreased number of parameters that can be dedicated to each language? Current evidence suggests slight interference effects, but also that identical models maintain much of their monolingual performance when trained on multilingual corpora \citep{mueller2020cross}. Is syntactic agreement, in particular, encoded independently per language or shared across languages? Some studies suggest that syntax is encoded in similar ways across languages \citep{chi2020mbert_probe,stanczak2022neurons}, though these rely on correlational methods based on dependency parsing, which introduce confounds and may not rely on syntactic information \emph{per se}.

\paragraph{Syntactic probing.} Various behavioral probing studies have analyzed the syntactic behavior of monolingual and multilingual LMs (\citealp{dupouxlinzen16,marvinlinzen18,ravfogel2019studying,mueller2020cross,hu2020systematic}). Results from behavioral analyses are generally easier to interpret and present clearer evidence for \emph{what} models' preferences are given various contexts. However, these methods do not tell us \emph{where} or \emph{how} syntax is encoded.

A parallel line of work employs parametric probes. Here, a linear classifier or multi-layer perceptron probe is trained to map from a model's hidden representations to dependency attachments and/or labels \citep{hewitt2019structural} to locate syntax-sensitive regions of a model. This approach has been applied in multilingual models \citep{chi2020mbert_probe}, and produced evidence for parallel dependency encodings across languages. However, if such probes are powerful, they may learn the target task themselves rather than tap into an ability of the underlying model \citep{hewitt2019selectivity}, leading to uninterpretable results. When controlling for this, even highly selective probes may not need access to syntactic information to achieve high structural probing performance \citep{sinha2021masked}. There are further confounds when analyzing individual neurons using correlational methods; for example, probes may locate encoded information that is not actually used by the model \citep{antverg2022pitfalls}.

\emph{Causal probing} has recently become more common for interpreting various phenomena in neural models of language. \citet{lakretz2019emergence} and \citet{LAKRETZ2021104699} search for syntax-sensitive units in English and Italian monolingual LSTMs by intervening directly on activations and evaluating syntactic agreement performance. \citet{vig2020causal} propose causal mediation analysis for locating neurons and attention heads implicated in gender bias in pre-trained language models; this method involves intervening directly on the inputs or on individual neurons. \citet{finlayson2021causal} extend this approach to implicate neurons in syntactic agreement. This study extends their data and method to multilingual stimuli and models.

Other causal probing work uses interventions on model representations, rather than inputs. This includes amnesic probing \citep{elazar2021amnesic}, where part-of-speech and dependency information is deleted from a model using iterative nullspace projection (INLP; \citealp{ravfogel2020null}). \citet{ravfogel2021counterfactual} employ INLP to understand how relative clause boundaries are encoded in BERT.

\section{Methods}
\subsection{Causal Metrics}
We first define terms to represent the quantities we measure before and after the intervention. We are interested in the impact of an intervention $\mathbf{x}$ on a model's preference $y_\mathbf{x}$ for grammatical inflections over ungrammatical ones. We start with the original input, on which we apply the \texttt{null} intervention: This represents performing no change to the original input. Given prompt $u$ and verb $v$, we first calculate the following ratio:
\begin{align}
    y_{\texttt{null}}(u,v) = \frac{p(v_{pl}\mid u_{sg})}{p(v_{sg}\mid u_{sg})}
\end{align}

Here, $u_{sg}$ represents a prompt that would require a singular verb inflection $v_{sg}$ at the \texttt{[MASK]} for the sentence to be grammatical; for example, ``The \textcolor{blue}{doctor} near the \textcolor{red}{cars} \texttt{[MASK]} it''. $v_{sg}$ is the third-person singular present inflection of verb $v$, and $v_{pl}$ is the plural present inflection; for example, $v_{sg}=$``\textcolor{blue}{observes}'' and $v_{pl}=$``\textcolor{red}{observe}''. Note that this ratio has the incorrect inflection as the numerator; this entails that if the model computes agreement correctly, we will have $y < 1$.

We now define the \texttt{swap-number} intervention, where the grammatical number of $u$ is flipped (resulting in ``The \textcolor{red}{doctors} near the \textcolor{red}{cars} \texttt{[MASK]} it'' for the previous example). This results in the following expression for $y$:

\begin{align}
    y_{\texttt{swap-number}}(u,v) = \frac{p(v_{pl}\mid u_{pl})}{p(v_{sg}\mid u_{pl})}
\end{align}
Now, the numerator is the correct inflection, so we expect $y > 1$.

As we are interested in the contribution of individual model components to the model's overall preference for correct inflections, we focus on \emph{indirect effects}, where we perform interventions on individual model components and observe the change in $y$. In particular, we measure the \textbf{natural indirect effect} (NIE), as follows.

We intervene on an individual neuron $\mathbf{z}$. We change \textbf{z}'s original activation given $u$ and $v$ (denoted $\mathbf{z}_{\texttt{null}}(u,v)$) to the activation it \emph{would have} taken if we had performed the intervention on $u$ (denoted $\mathbf{z}_{\texttt{swap-number}}(u,v)$). The rest of the neurons retain their original activations. ``Natural'' here refers to the fact that our intervention changes the activation $\mathbf{z}$ to the value it would have in another natural setting $u'$, rather than setting it to some predefined constant (such as 0) that it may or may not obtain given natural inputs. We measure the relative change in $y$ after applying the intervention (see Figure~\ref{fig:nie} for a visual example):

\begin{figure}
    \centering
    \includegraphics[width=\linewidth]{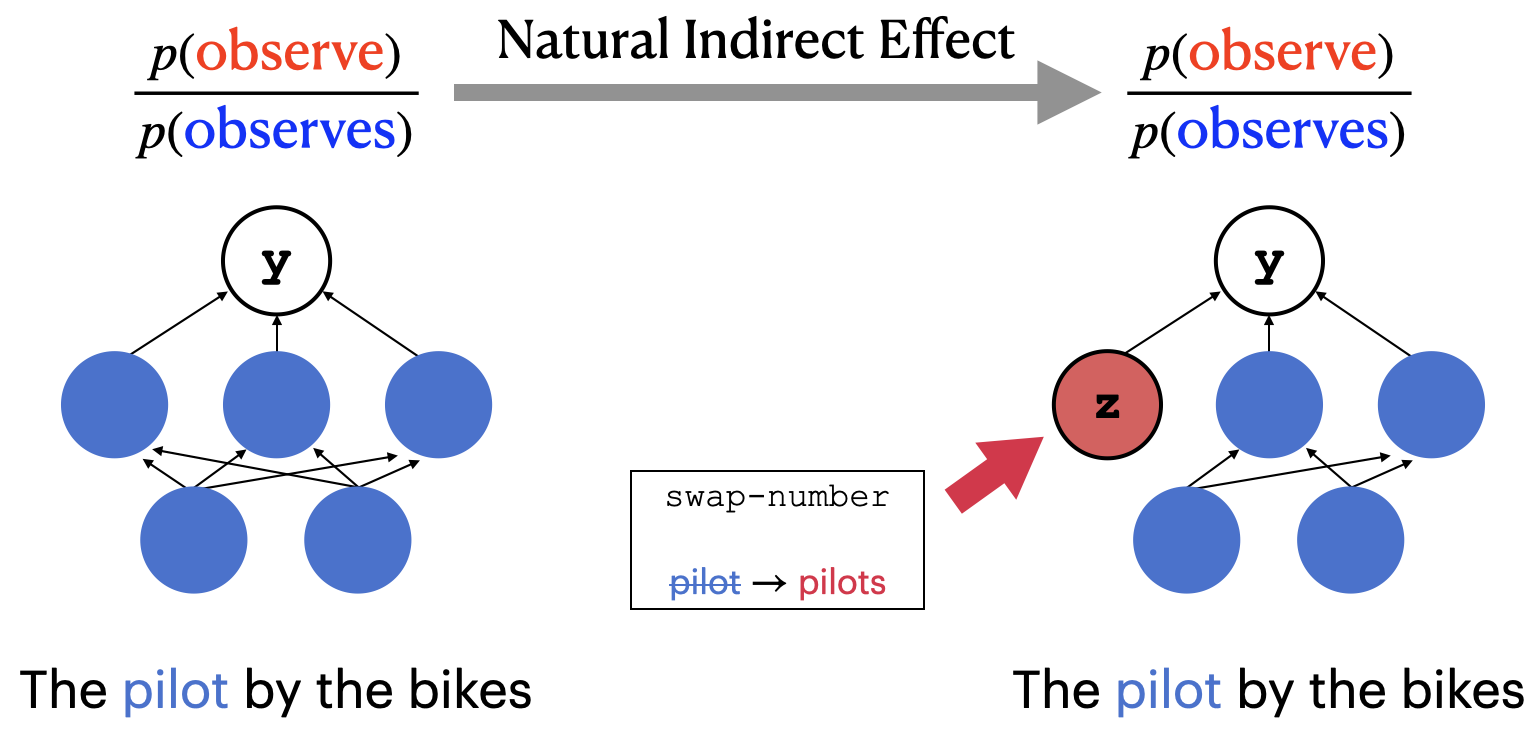}
    \caption{Example of computing the natural indirect effect (NIE). We change a neuron's activation to what it would have been if we had intervened on the prompt, then measure the relative change in $y$.}
    \label{fig:nie}
\end{figure}

\begin{align}\label{eq:nie}
    \begin{split}
    \overline{\text{NIE}}(\texttt{swap-number, null};y,\mathbf{z}) = \\
    \mathbb{E}_{u,v}\left[ \frac{y_{\texttt{null},\mathbf{z}_{\texttt{swap-number}}(u,v)}(u,v) - y_{\texttt{null}}(u,v)}{y_\texttt{null}(u,v)}\right] = \\
    \mathbb{E}_{u,v}\left[\frac{y_{\texttt{null},\mathbf{z}_{\texttt{swap-number}}(u,v)}(u,v)}{y_\texttt{null}(u,v)} - 1\right]
    \end{split}
\end{align}

If a neuron encodes useful information for syntactic agreement, we expect $y$ to \emph{increase} after the intervention, making the numerator positive. Positive NIEs indicate that a neuron encodes preferences for correct verb inflections, and negative NIEs indicate that the neuron prefers incorrect inflections. The closer the NIE is to 0, the less of a contribution a neuron makes to syntactic agreement in either direction.

\subsection{Models}
\citet{finlayson2021causal} analyzed a series of monolingual autoregressive language models (ALMs): GPT-2 \citep{gpt2}, TransformerXL \citep{transformerxl}, and XLNet \citep{xlnet}. Here, we apply their analysis approach to multilingual models. Multilingual ALMs are rare in the literature; to our knowledge, the only ALM designed to be multilingual is XGLM \citep{xglm},\footnote{GPT-3 \citep{gpt3} is technically multilingual, as its training corpus contains data from other languages. However, it was not designed with multilinguality in mind, and the vast majority of its training data is English.} which we employ in this study.

Multilingual MLMs are much more common. We focus on multilingual BERT \citep{bert}. We were unable to analyze XLM-R \citep{xlmr}, a more recent multilingual MLM that performs better than mBERT on certain benchmarks, since its tokenizer splits a large proportion of our nouns and verbs into multiple tokens, which greatly constrained the stimuli we could use. In future work, we intend to address this issue by developing methods that enable multi-token interventions, as well as calibrated comparisons across variable-length sequences.

\begin{table}[t]
    \centering
    \resizebox{0.8\linewidth}{!}{
    \begin{tabular}{lrrr}
    \toprule
    Model & Layers & Neurons & Parameters \\
    \midrule
    BERT & 12 & 9984 & 110M \\
    mBERT & 12 & 9984 & 110M \\
    \midrule
    GPT-2 & 24 & 25600 & 345M \\
    XGLM & 24 & 25600 & 564M \\
    \bottomrule
    \end{tabular}}
    \caption{The size of each model used in this study. Each monolingual BERT variant (including the RoBERTa-based CamemBERT) has the same number of layers, neurons, and parameters as BERT.}
    \label{tab:models}
\end{table}

We also analyze a series of monolingual MLMs---one for each language included in our sample. Four of these models were based on BERT: BERT (English), GermanBERT,\footnote{\url{https://www.deepset.ai/german-bert}} BERTje (Dutch; \citealp{bertje}), and FinnishBERT \citep{bert-finnish}. The French MLM, CamemBERT \citep{camembert}, is based on RoBERTa \citep{roberta}, which is architecturally near-identical to BERT.

\subsection{Materials}\label{sec:data}
\newlength{\vs}
\setlength{\vs}{0.4\baselineskip}
\begin{figure}[t]
        \rule{\columnwidth}{1pt}
        \centering
        \resizebox{0.95\columnwidth}{!}{
        
        \begin{minipage}{\columnwidth}
            \raggedright
            \vspace{0.2cm}
         \textit{Simple Agreement}:\\
         The \textcolor{blue}{athlete} \underline{\textcolor{blue}{investigates}/*\textcolor{red}{investigate}}\ldots
        
        \rule{0.9\columnwidth}{0.6pt}
         
        \vspace{\vs}\textit{Across Prepositional Phrase}:\\
        The \textcolor{blue}{manager} behind the \textcolor{red}{bikes}
        \underline{\textcolor{blue}{observes}/*\textcolor{red}{observe}}\ldots

        \vspace{\vs}\textit{Across Object Relative Clause:}\\
        The \textcolor{blue}{farmers} that the \textcolor{red}{parent} loves \underline{*\textcolor{red}{confuses}/\textcolor{blue}{confuse}}\ldots
        \vspace{0.2cm}
        \end{minipage}
    }
        \rule{\columnwidth}{1pt}
    \caption{Constructions used in this study, grouped by whether the subject and verb are adjacent. We use a subset of constructions from \citet{finlayson2021causal}, directly translating the stimuli to French, German, Dutch, and Finnish. See Appendix~\ref{sec:examples} for examples of each structure in each language.}
    \label{fig:examples}
\end{figure}

\begin{figure*}
    \centering
    \includegraphics[width=0.75\linewidth]{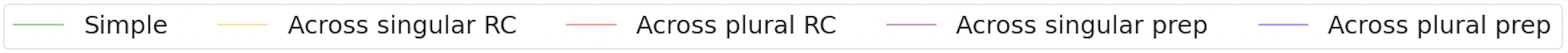}
    
    \begin{subfigure}{0.19\linewidth}
        \includegraphics[height=2.5cm,trim={0 0 14.4cm 0},clip]{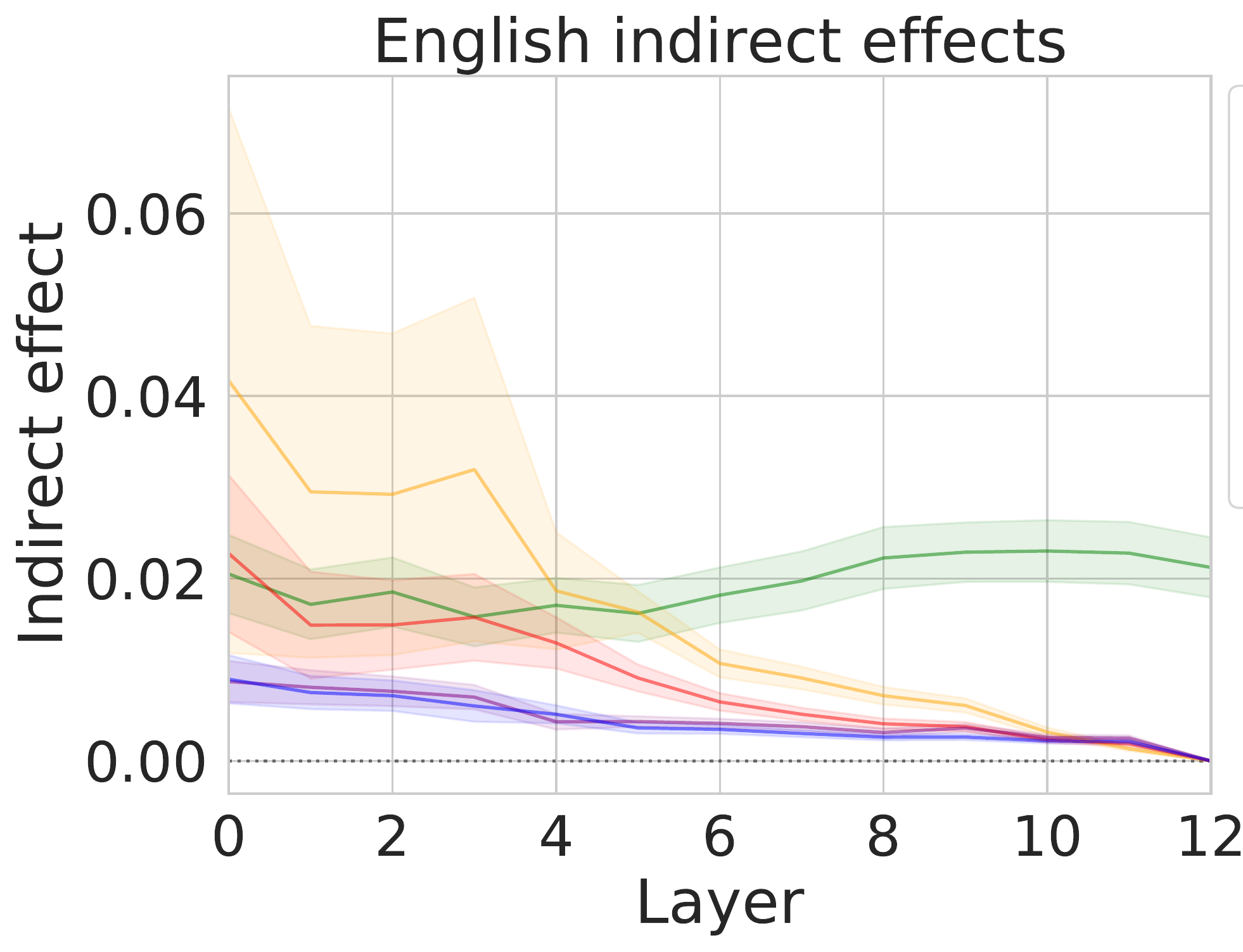}
        \caption{BERT}
    \end{subfigure}
    \hfill
    \begin{subfigure}{0.19\linewidth}
        \includegraphics[height=2.5cm,trim={1.25cm 0 13.8cm 0},clip]{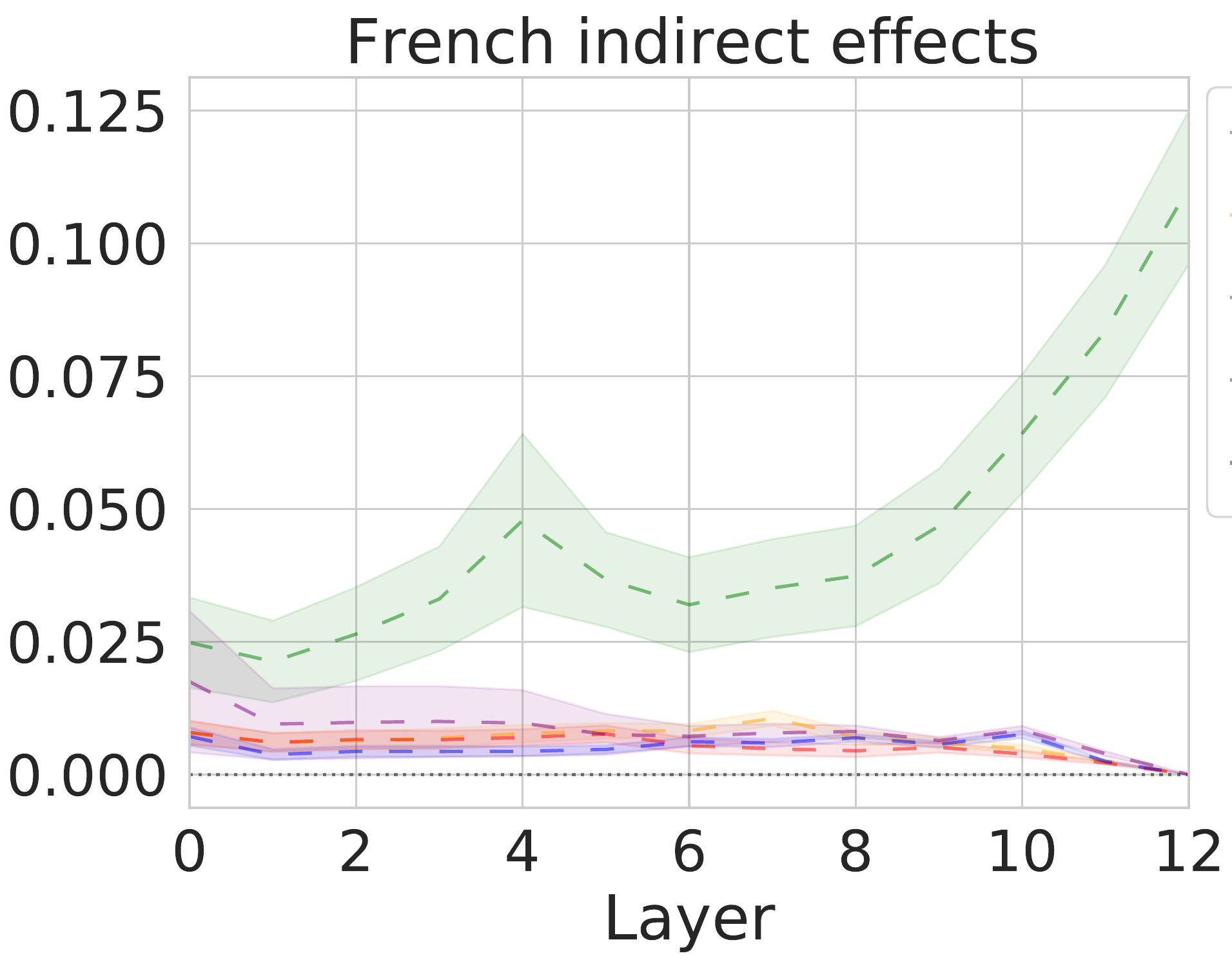}
        \caption{CamemBERT}
    \end{subfigure}
    \hfill
    \begin{subfigure}{0.19\linewidth}
        \includegraphics[height=2.5cm,trim={1.25cm 0 13.8cm 0},clip]{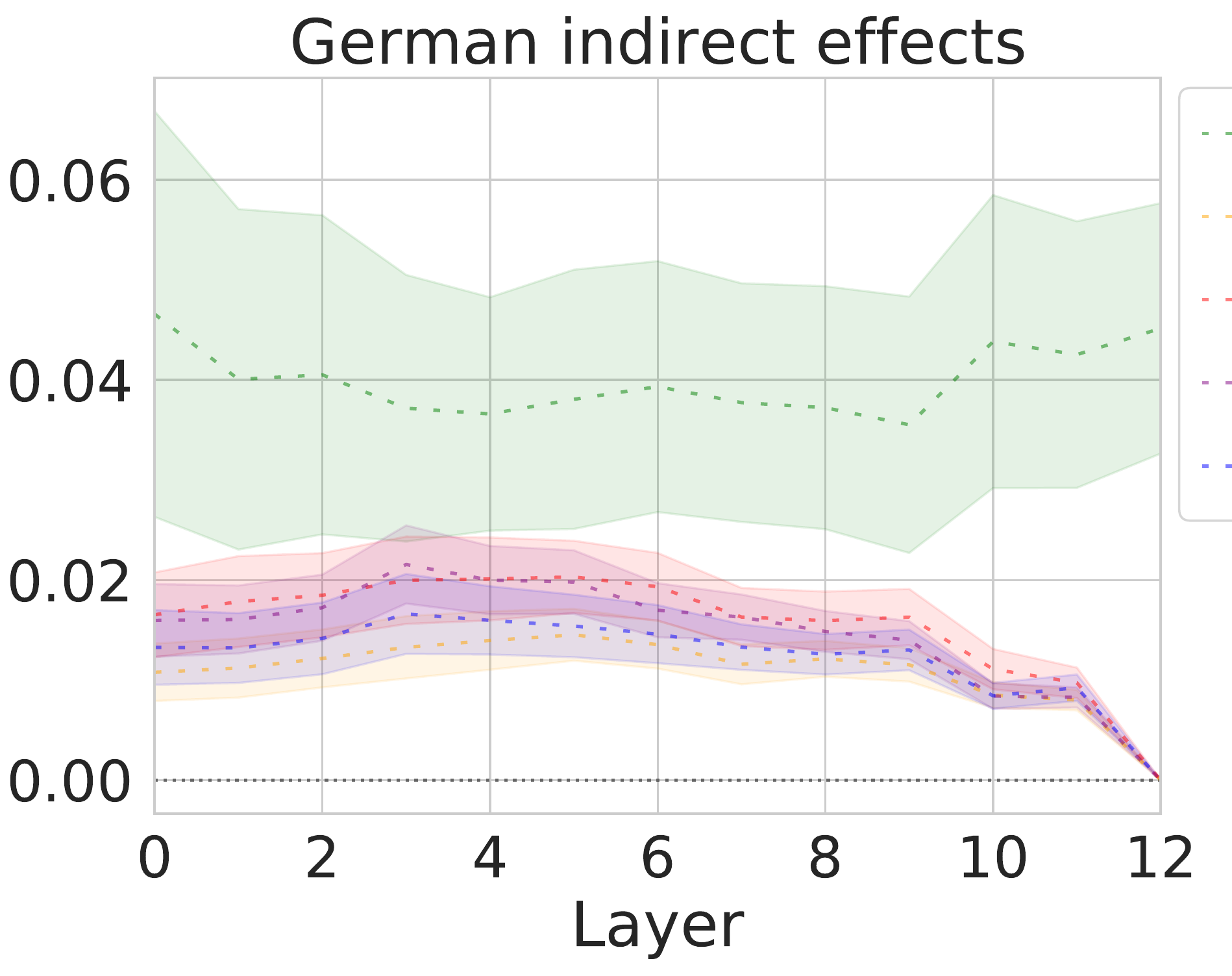}
        \caption{GermanBERT}
    \end{subfigure}
    \begin{subfigure}{0.19\linewidth}
        \includegraphics[height=2.5cm,trim={1.25cm 0 13.8cm 0},clip]{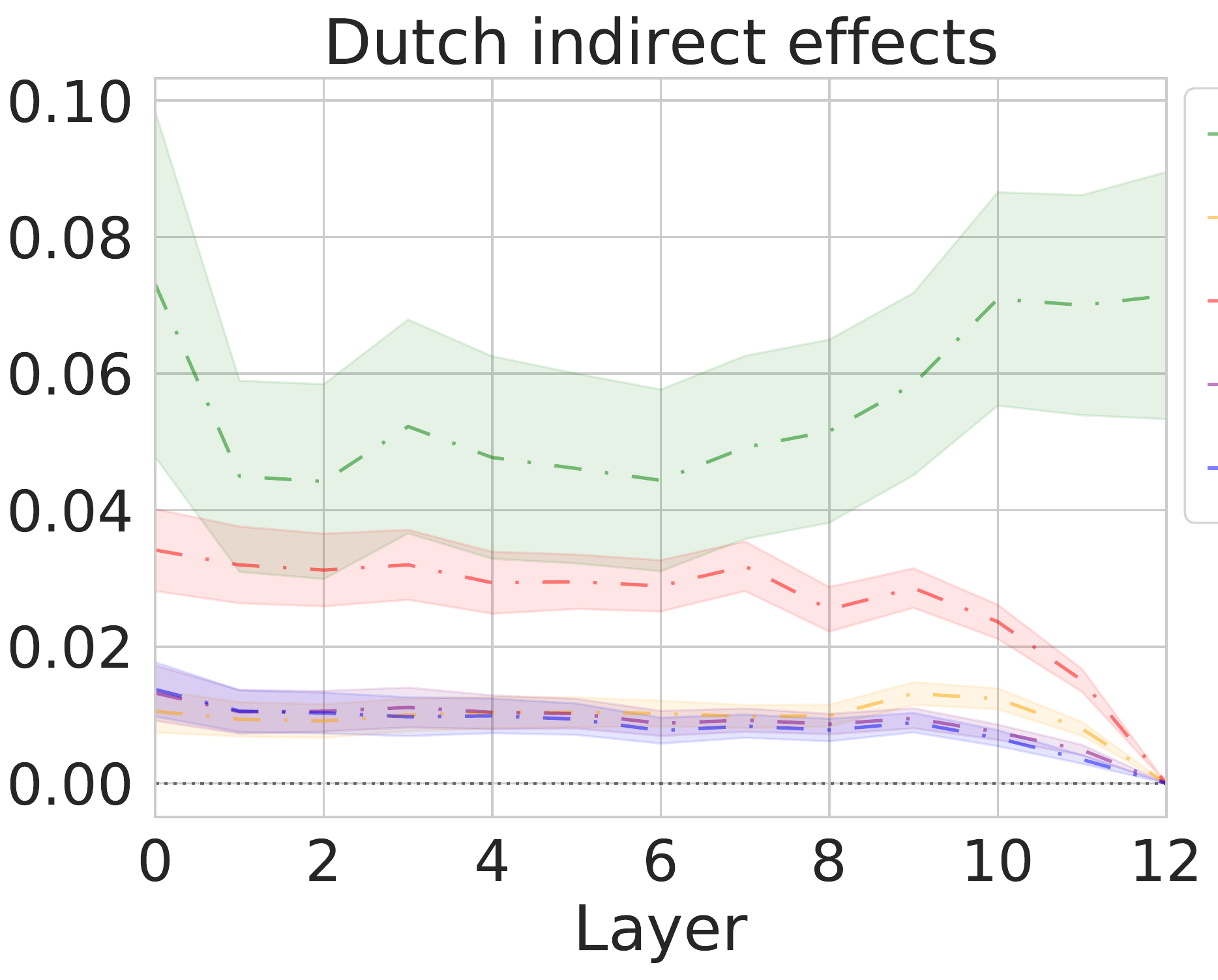}
        \caption{BERTje}
    \end{subfigure}
    \begin{subfigure}{0.19\linewidth}
        \includegraphics[height=2.5cm,trim={1.25cm 0 13.8cm 0},clip]{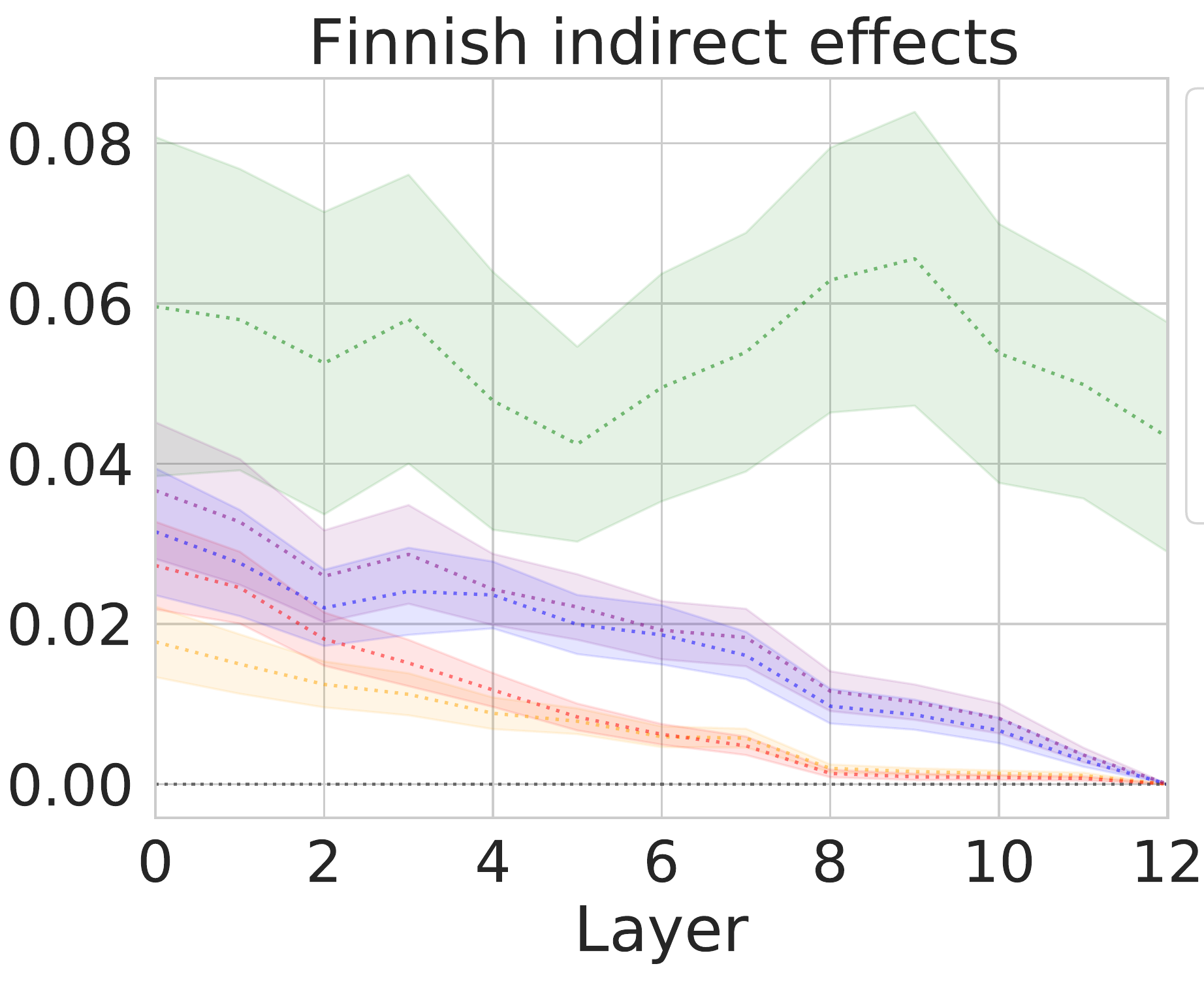}
        \caption{FinnishBERT}
    \end{subfigure}
    \caption{Natural indirect effects for the top 5\% of neurons in each layer for monolingual masked language models. There are two distinct layer-wise NIE contours in each language, depending on whether the subject and verb are separated by other tokens (as in `across a relative clause' and `across a prepositional phrase' structures) or not (as in `simple agreement').}
    \label{fig:indirect_effects_mlm}
\end{figure*}

We translate the stimuli from \citet{finlayson2021causal} (Figure~\ref{fig:examples}) to French, German, Dutch, and Finnish. Since the subjects and verbs on which we intervene must be one token each,\footnote{It is not clear how to compare the probability of variable-length sequences in masked language models, and autoregressive language models tend to prefer sequences containing fewer tokens. There have been attempts to compare variable-length sequence probabilities using iterative approaches \citep[e.g.,][]{schick2021exploiting}, though this generally requires fine-tuning to work properly.} we are restricted to very frequent words in the pre-training corpus which do not get split into subwords by a model's tokenizer. This limits us to high-resource language varieties---and as most of the top languages in mBERT and XGLM's pre-training corpora are Indo-European, this also limits the typological range of this method. A virtue of our sample of languages, however, is that is allows us to study whether neurons are shared across typologically similar languages, where shared neurons and similar layer-wise effect patterns are most likely to occur: If syntactic agreement neurons are not shared across similar languages, they are unlikely to be shared across \emph{any} languages.

For each structure, we sample up to 200 sentences. If there are fewer than 200 sentences where the subjects and verbs are single tokens, we take the entire set of valid stimuli. When we use the original stimuli from \citet{finlayson2021causal}, we often have very few sentences where the subjects and verbs are single tokens. Thus, we also create short-word versions of the stimuli, where we use shorter and more common words (e.g., instead of "managers" or "observe", we can use "cats" or "see"). Our results are consistent when using the original and short nouns and verbs; see Appendix~\ref{app:short_long_stimuli}.

\begin{figure}[!t]
    \centering
    \includegraphics[width=0.88\linewidth]{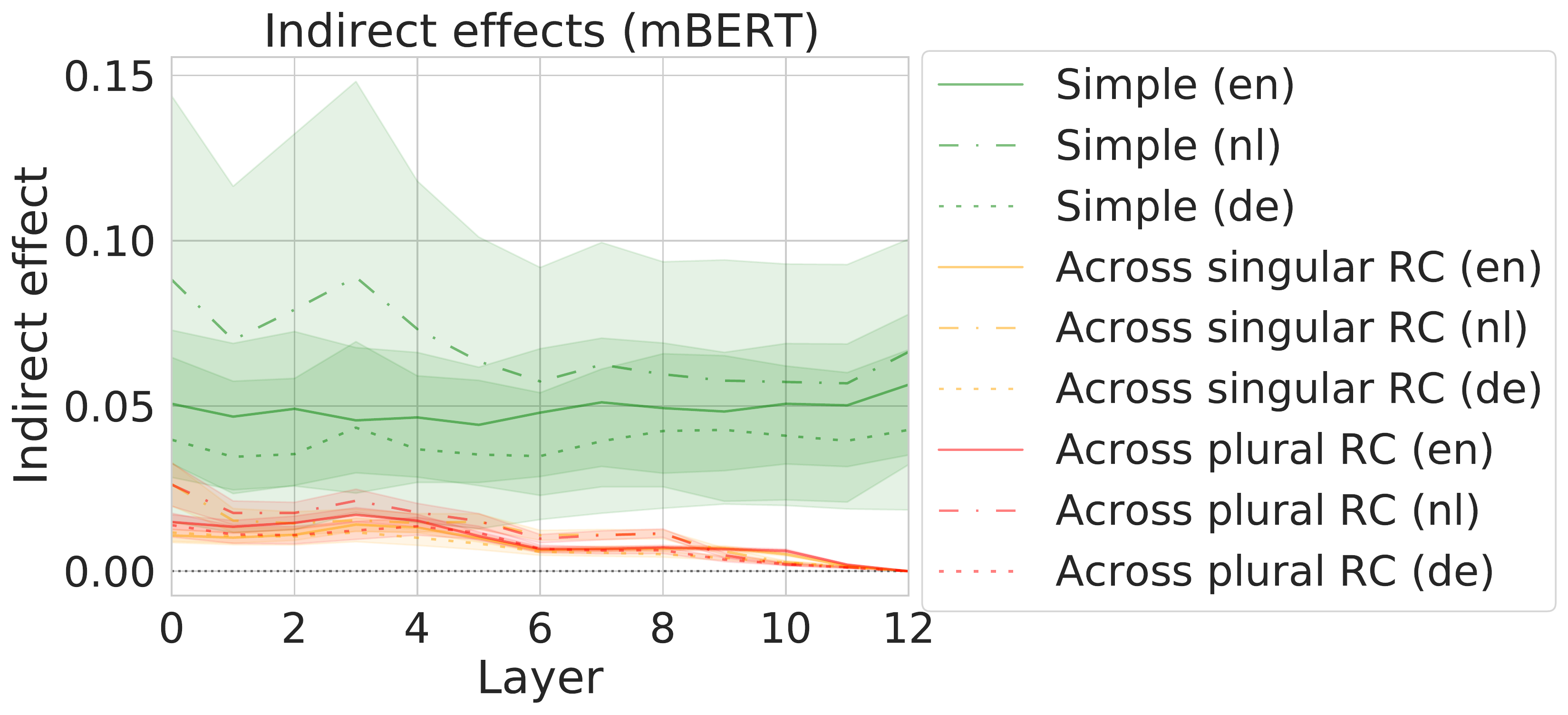}
    \includegraphics[width=0.88\linewidth]{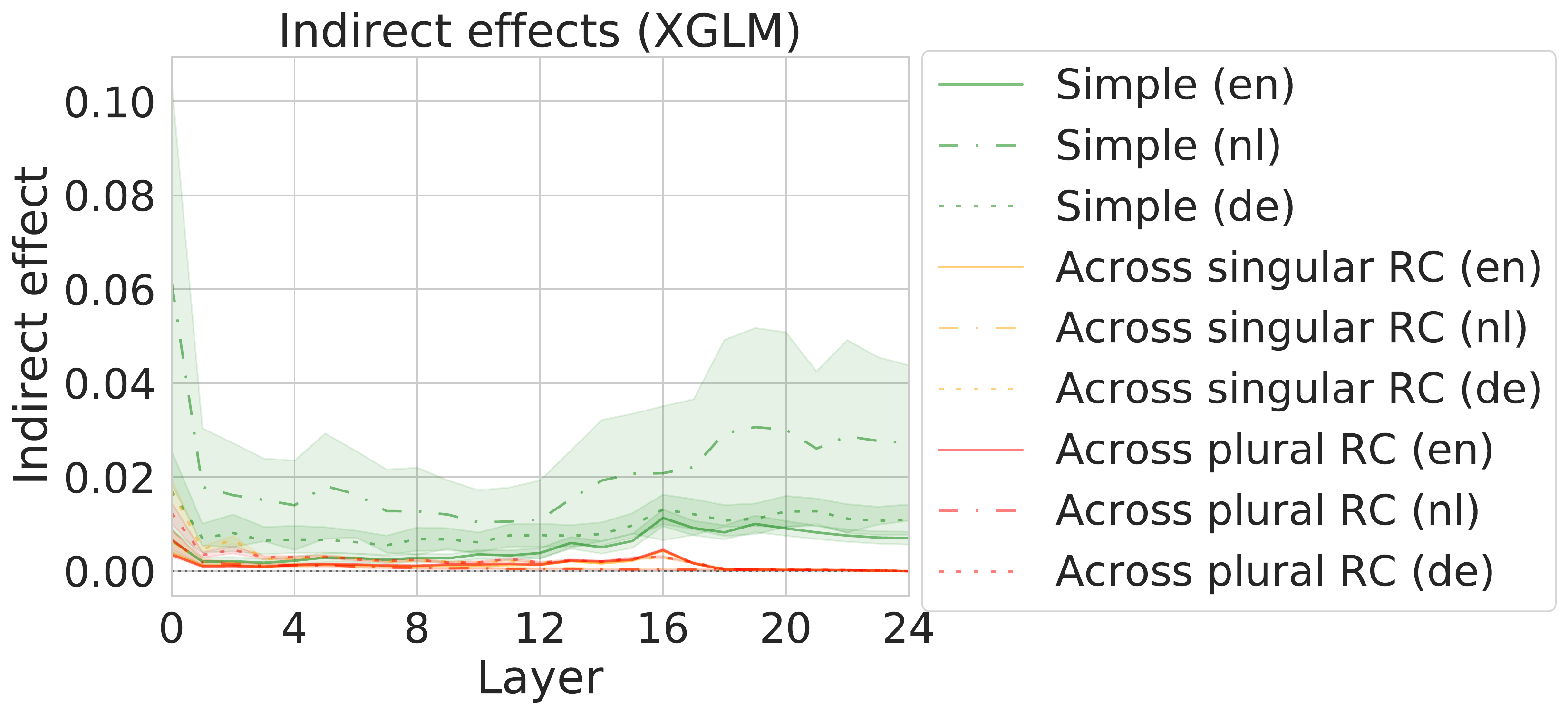}
    \caption{Natural indirect effects for mBERT (top) and XGLM (bottom) for Germanic languages. There are two distinct layer-wise NIE patterns in each language. NIE patterns for the same structure look very similar across languages.}
    \label{fig:indirect_effects_mbert_xglm}
\end{figure}

The original stimuli were generated from a grammar given a list of manually selected terminals. By generating artificial stimuli and \emph{not} sampling sentences from a corpus, we partially control for memorized sequences or token collocations in the pre-training corpus.

\citet{finlayson2021causal} found two distinct layer-wise NIE patterns for syntactic agreement: one when the subject and verb are adjacent (the \emph{short-range} effect), and another when they are separated by any number of tokens (the \emph{long-range} effect). To understand whether the short-range effect is due to preferences for frequent bigrams (rather than specifically grammatical subject-verb bigrams), we also design a bigram swap intervention. We use high-mutual-information adjective-noun English bigrams as the original inputs and intervene by randomly swapping the first or second words in the bigram with words from a different bigram.
For example, given the bigrams \textcolor{blue}{coaxial cable} and \textcolor{red}{police officer}, we can define $y_{\texttt{null}} = \frac{p(\text{\textcolor{red}{officer}} \mid \text{\textcolor{blue}{coaxial}})}{p(\text{\textcolor{blue}{cable}} \mid \text{\textcolor{blue}{coaxial}})}$ and $y_{\texttt{swap-bigram}} = \frac{p(\text{\textcolor{red}{officer}} \mid \text{\textcolor{red}{police}})}{p(\text{\textcolor{blue}{cable}} \mid \text{\textcolor{red}{police}})}$. Then we can compute the NIE as in Equation~\ref{eq:nie}. 

Finally, to test whether separate neurons are used for short- and long-range token collocations in general, we also define short- and long-range \emph{semantic plausibility} baselines, where nouns are associated with stereotypical adjectives (e.g., \textcolor{blue}{square T.V.} and \textcolor{red}{red apple}). The short-range semantic plausibility intervention is the same as for the bigram intervention: We compute the probability ratio of the first and second noun in a pair of bigrams before and after swapping the adjective. For long-range semantic plausibility, the prompt $u$ is ``The \textcolor{blue}{T.V.}/\textcolor{red}{apple} is'', and $v$ is the probability ratio of the \emph{adjectives} before and after swapping the nouns.

\section{Results}
\subsection{Layer-wise NIE contours are similar across languages}\label{sec:results_nie}

We present indirect effects for monolingual masked language models (Figure~\ref{fig:indirect_effects_mlm}), as well as mBERT and XGLM (Figure~\ref{fig:indirect_effects_mbert_xglm}). Here, we select the top 5\% of neurons per layer by NIE. In each language, whether in a monolingual or multilingual MLM or ALM, \textbf{there are two distinct layer-wise NIE effect patterns for number agreement}: one for short-range dependencies and one for long-range dependencies. This agrees with the findings of \citet{finlayson2021causal} on autoregressive English LMs. However, these effects look more distinct across monolingual models, whereas \textbf{multilingual models exhibit more similar layer-wise NIE patterns across languages.} In other words, monoligual models accomplish syntactic agreement in different layers and neurons depending on the language (even though these languages are typologically similar), but in multilingual models agreement computations implicate the same layers across languages. This does not necessarily mean that the same individual neuron are being used cross-linguistically in multilingual models (we explore this question in more detail in \S\ref{sec:overlap_lang}); rather, the model may simply be learning similar layer-wise strategies for each language.

While prior work finds that syntactic agreement is easier to learn in languages that have more explicit morphological cues to hierarchical structure \citep{ravfogel2019studying,mueller2020cross},\footnote{Explicit case marking correlates well with performance on syntactic evaluations \citep{ravfogel2019studying}, so we would expect German and Finnish to exhibit different results if these cues give rise to different agreement computations.} this does not necessarily imply that different agreement mechanisms are learned in such languages. We find similar layer-wise NIEs in mBERT across each language we consider, including Finnish, a non-Indo-European (specifically, Uralic) language.

\subsection{Syntax neurons are shared across structures, but not with semantic baselines}\label{sec:semantic_baselines}
Here, we analyze to what extent the same neurons are implicated across syntactic structures and languages in mBERT. For each structure, we take the top 30 neurons by indirect effect (from any layer); we then compute the proportion of such high-NIE neurons that are shared across structures.

\begin{figure}[!t]
    \centering
    \begin{subfigure}{0.9\linewidth}
        \includegraphics[width=0.9\linewidth]{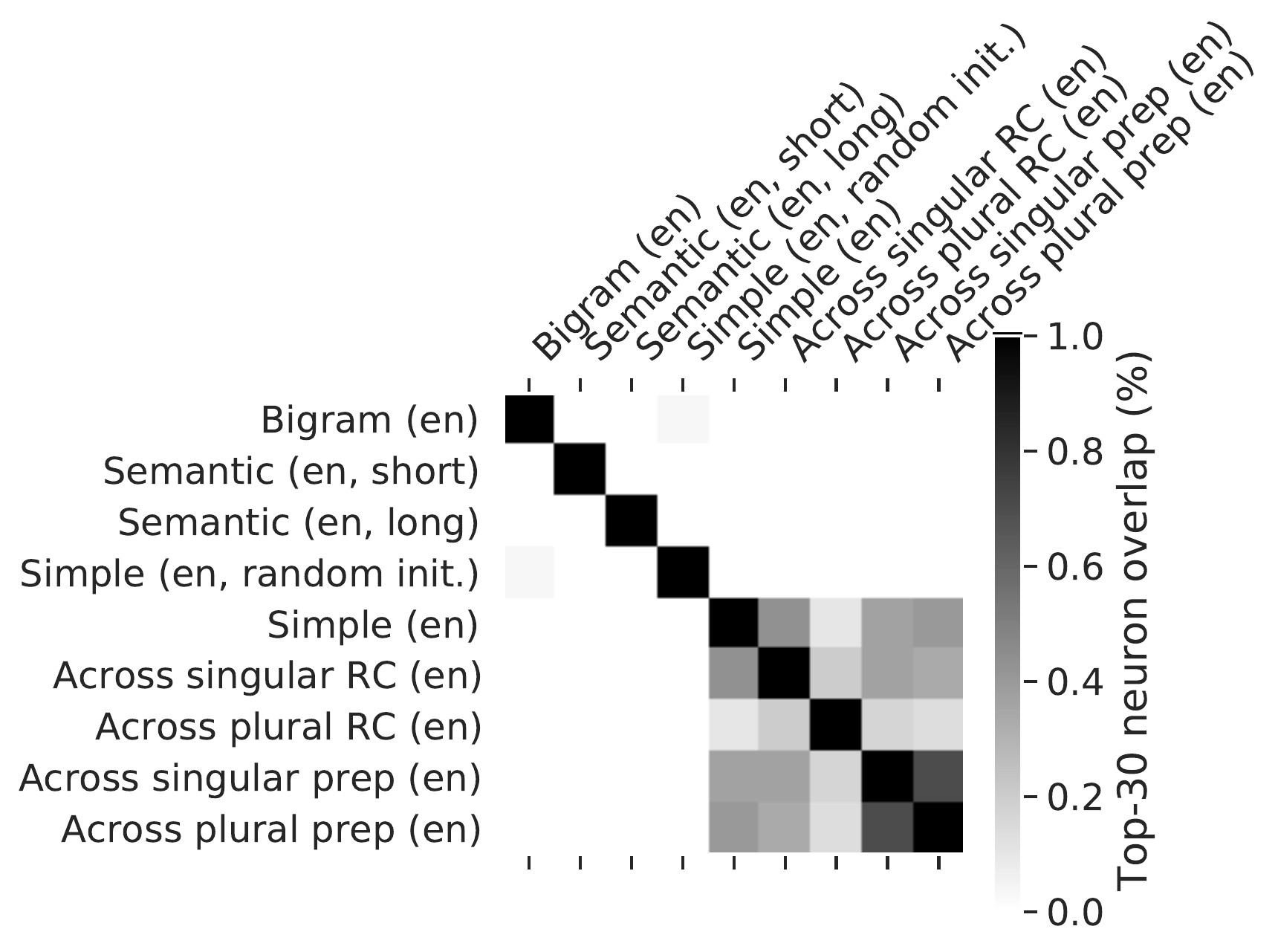}
        \caption{mBERT}
    \end{subfigure}
    \begin{subfigure}{0.9\linewidth}
        \includegraphics[width=0.9\linewidth]{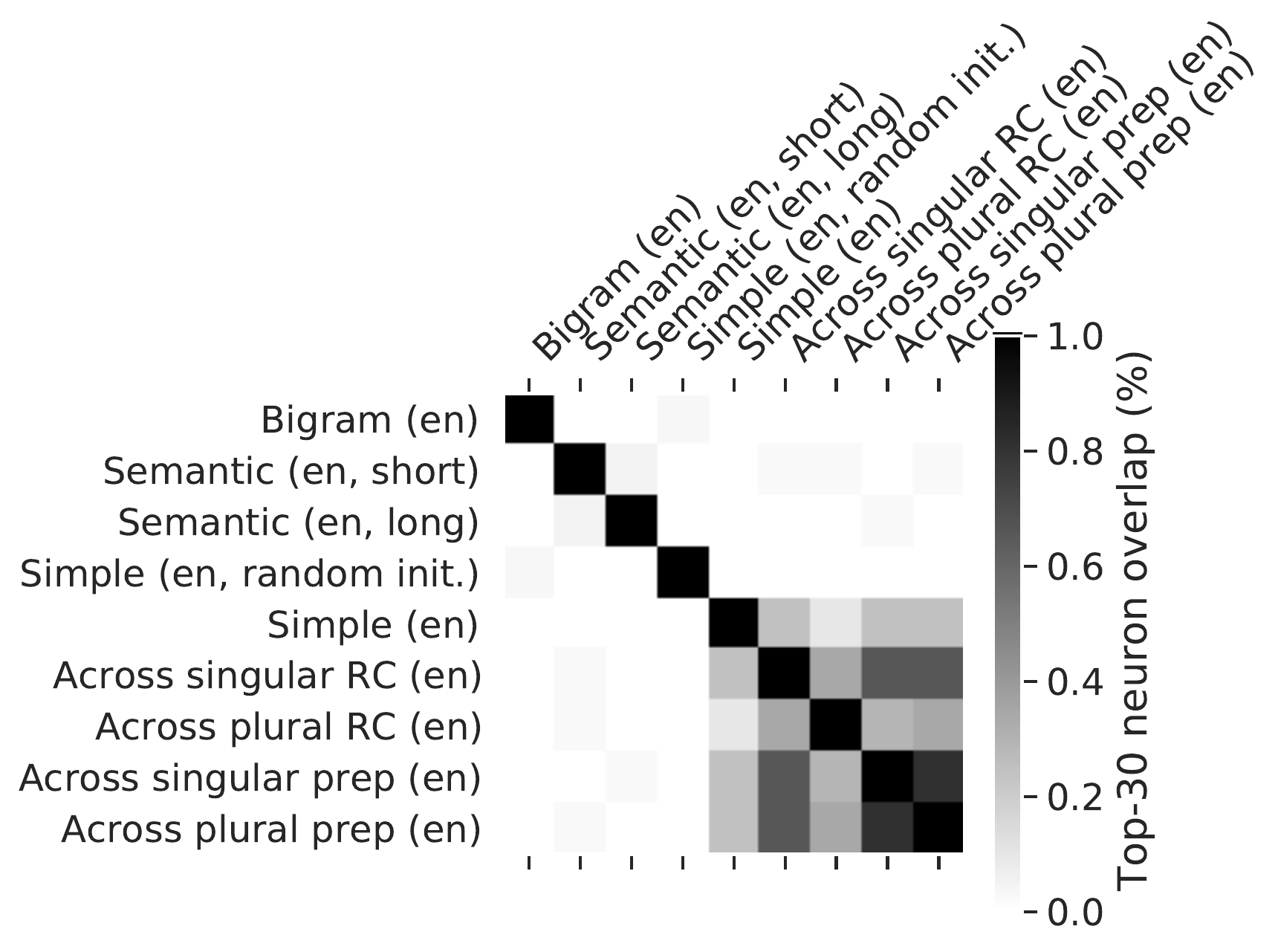}
        \caption{XGLM}
    \end{subfigure}
    \caption{Neuron overlap across structures (including baselines) in English for (a) mBERT and (b) XGLM. There is zero or near-zero overlap between the baselines and all syntactic agreement structures, whereas overlap is relatively high (and statistically significant) for all other structures.}
    \label{fig:overlap_en}
\end{figure}

First, we investigate to what extent the neurons that have high NIE for the syntactic structures are selective to syntax. We do so by computing the overlaps in English between neurons with high NIE for syntactic structures and the neurons with high NIE for our bigram and semantic plausibility baselines. We find that the top syntactic agreement neurons for any structure are \emph{not} shared with the neurons implicated in semantic plausibility or bigram collocation (Figure~\ref{fig:overlap_en}). In other words, \textbf{the neurons used for syntactic agreement are \emph{specific to agreement}} and do not track common bigrams more generally. 

Figure~\ref{fig:overlap_en} also shows that \textbf{neurons are shared across syntactic structures}, providing evidence for an abstract notion of syntactic agreement encoded in mBERT that is separate from the individual structures that the model is presented with. However, the varying extents of overlap indicate that there are also neurons specialized to particular structures. To further contextualize these overlap proportions, we also compute overlaps for simple agreement in a randomly initialized mBERT, as a baseline. This experiment yields near-zero overlaps, indicating that the overlaps across structures we obtain for mBERT and XGLM are unlikely to be due to random chance.\footnote{For reference, the probability of at least one neuron being shared between two random samples of 30 neurons in (m)BERT-base is $1 - \frac{{9984-30 \choose 30}}{{9984 \choose 30}} \approx .086$.}

\subsubsection{Neurons are shared across languages in autoregressive language models}\label{sec:overlap_lang}
\begin{figure}[!t]
    \centering
    \begin{subfigure}{0.9\linewidth}
        \includegraphics[width=0.9\linewidth]{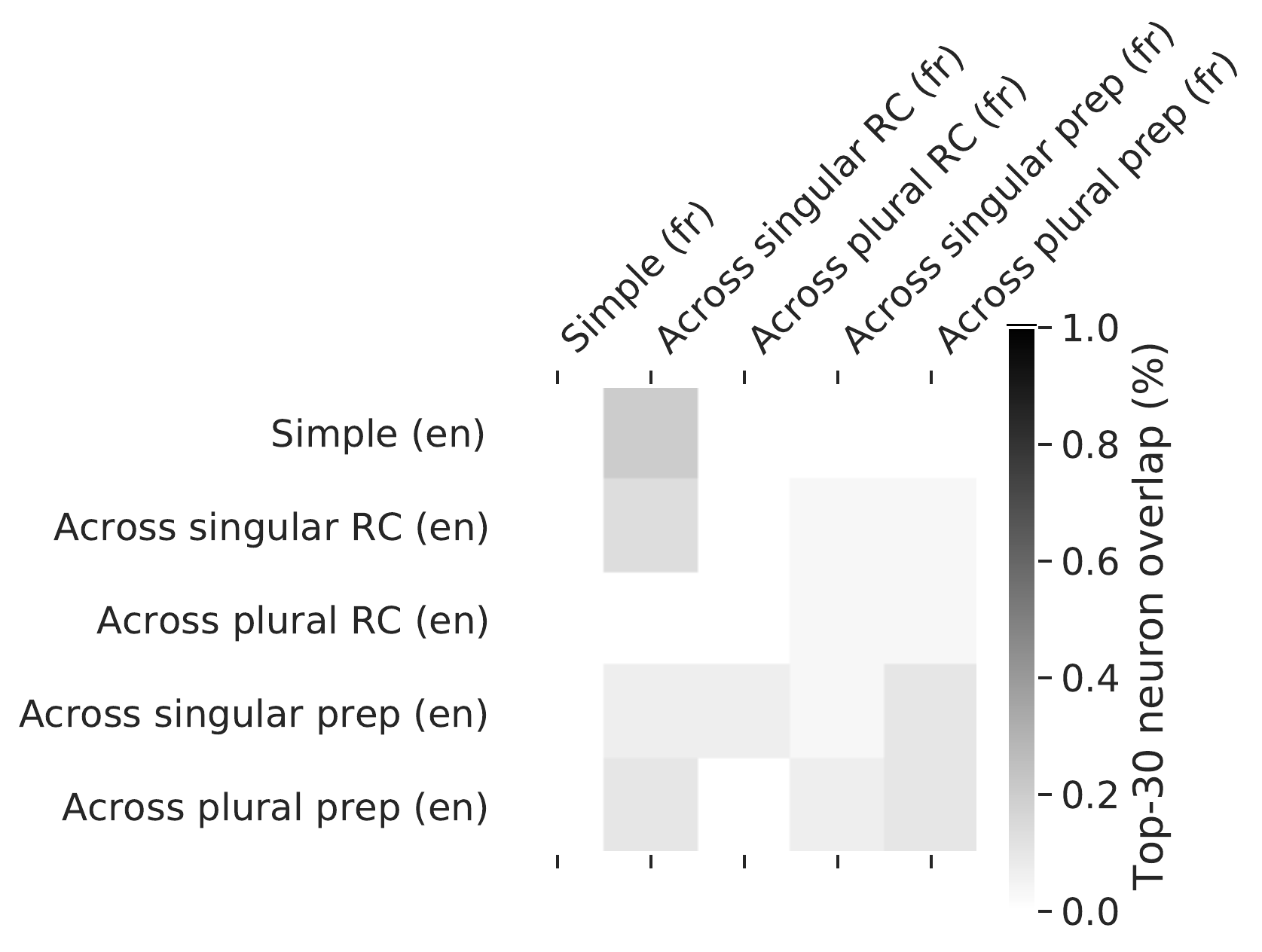}
        \caption{mBERT}
        \label{fig:overlap_en_de}
    \end{subfigure}
    \begin{subfigure}{0.9\linewidth}
    \centering
        \includegraphics[width=0.9\linewidth]{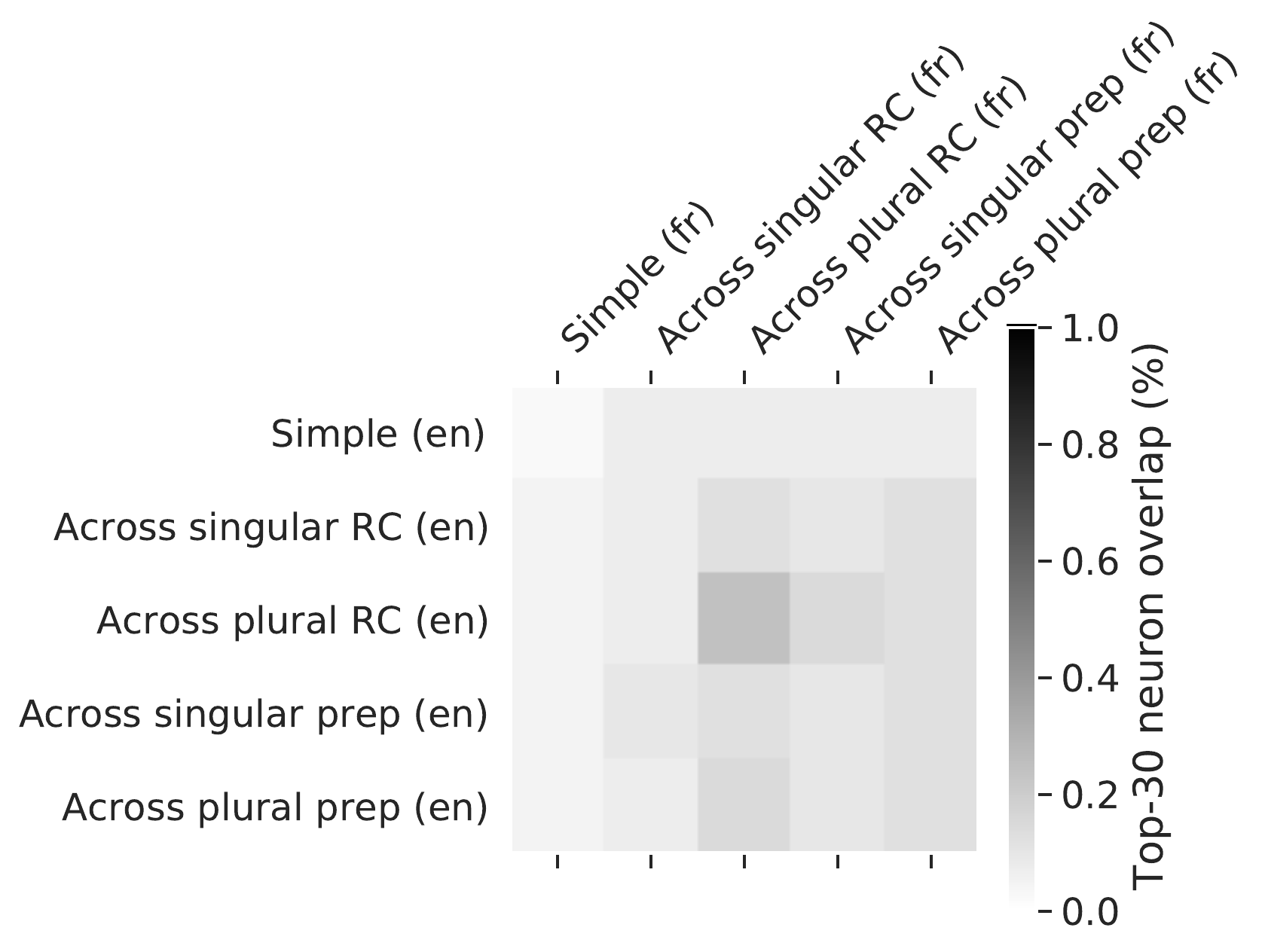}
        \caption{XGLM}
    \end{subfigure}
    \caption{Cross-lingual neuron overlaps for the top 30 neurons by NIE in (a) mBERT and (b) XGLM. We present English-French overlaps; overlaps between other language pairs look similar (see Appendix~\ref{app:neuron_overlap_all}). The overlap percentages in (b) are significantly higher than random chance. Overlaps for most structure pairs in (a) are not significant, except for overlaps between `Across a preposition' structures and other long-range agreement structures.}
    \label{fig:overlap}
\end{figure}

The overlap in neurons across languages (Figure~\ref{fig:overlap}) is significant for all structures in XGLM. For mBERT, overlap is significant between ``across a PP'' structures and other long-distance agreement structures, but not for any other structure pairs. Note that in XGLM, the diagonal is no darker than most other squares; in other words, there is \emph{not} more cross-lingual neuron overlap for the same syntactic structure relative to other structures. \textbf{These may be generic cross-lingual syntax neurons which are not specialized to any particular structure or language.} We found in \S\ref{sec:semantic_baselines} that there is almost no overlap between syntactic agreement neurons and bigram collocation/semantic plausibility neurons in English, which is further evidence that these may be more general syntactic agreement neurons. Nonetheless, overlap is very low across languages compared to across structures within a given language. Thus, \textbf{in autoregressive language models, syntactic agreement neurons can be language-specific or cross-lingual, but most are language-specific. For masked language models, syntactic agreement neurons are rarely shared across languages.}

\subsection{Neuron sparsity differs across structures, but not across languages}\label{sec:sparsity}

\begin{table}[t]
    \centering
    \resizebox{\linewidth}{!}{
    \begin{tabular}{llrr}
    \toprule
    Language & Model & \% Neurons for TE & \% Neurons for Max.\ NIE \\
    \midrule
    \multirow{4}{*}{\texttt{en}} & BERT & 1.0\% & 5.8\% \\
    & mBERT & 1.0\% & 8.7\% \\
    & GPT-2 & 17.5\% & 25.0\% \\
    & XGLM & 4.5\% & 16.5\% \\
    \midrule
    \multirow{3}{*}{\texttt{fr}} & CamemBERT & 6.7\% & 10.6\% \\
    & mBERT & 3.8\% & 29.8\% \\
    & XGLM & 3.5\% & 24.0\% \\
    \midrule
    \multirow{3}{*}{\texttt{de}} & GermanBERT & 1.0\% & 8.7\% \\
    & mBERT & 1.0\% & 6.7\% \\
    & XGLM & 1.5\% & 18.0\% \\
    \midrule
    \multirow{3}{*}{\texttt{nl}} & BERTje & 1.0\% & 5.8\% \\
    & mBERT & 1.0\% & 2.9\% \\
    & XGLM & 0.5\% & 37.5\% \\
    \midrule
    \texttt{fi} & FinnishBERT & 1.0\% & 4.8\% \\
    \bottomrule
    \end{tabular}}
    \caption{Neuron sparsities for the ``simple agreement'' structure across languages and models. Multilingual models do not necessarily encode syntax more sparsely than monolingual models. Sparsities are generally consistent across languages for the same model.}
    \label{tab:sparsity}
\end{table}

What proportion of LMs' neurons encode subject-verb agreement? The sparsity of syntax neurons in pre-trained models may vary depending on which language and structure we observe. \citet{lakretz2019emergence} and \citet{LAKRETZ2021104699} found that agreement neurons are sparse in LSTMs, but it is not clear whether this would hold for MLMs or large Transformer-based ALMs. Given our consistent results across languages, we hypothesize that the neuron sparsity of subject-verb agreement will be similar across monolingual models. Given the consistent distinction thus far in how neurons encode short- and long-range agreement, we also hypothesize that neuron sparsity will differ between agreement distances. Due to lower parameterization per language in multilingual models, however, we hypothesize that multilingual models encode agreement more sparsely than monolingual models.

We measure sparsity by iteratively selecting the top $k$ neurons by NIE, intervening on them simultaneously, and computing the natural indirect effect after performing the \texttt{swap-number} intervention. We continue sampling $k$ more neurons and computing NIEs until we have selected all neurons; the NIE after intervening on all neurons is equivalent to the \emph{total effect} (TE).\footnote{Intuitively, the TE can be thought of as the preference of the model as a whole for correct verbs over incorrect verbs.} Computing effects for each neuron and each structure is computationally expensive, so we use $k=128$ for XGLM and GPT-2 (0.5\% of neurons selected at a time) and $k=96$ for (m)BERT ($\approx$1.0\% of neurons selected at a time).

\begin{figure}[!t]
    \centering
    \begin{subfigure}{0.99\linewidth}
        \centering
        \includegraphics[width=0.7\linewidth]{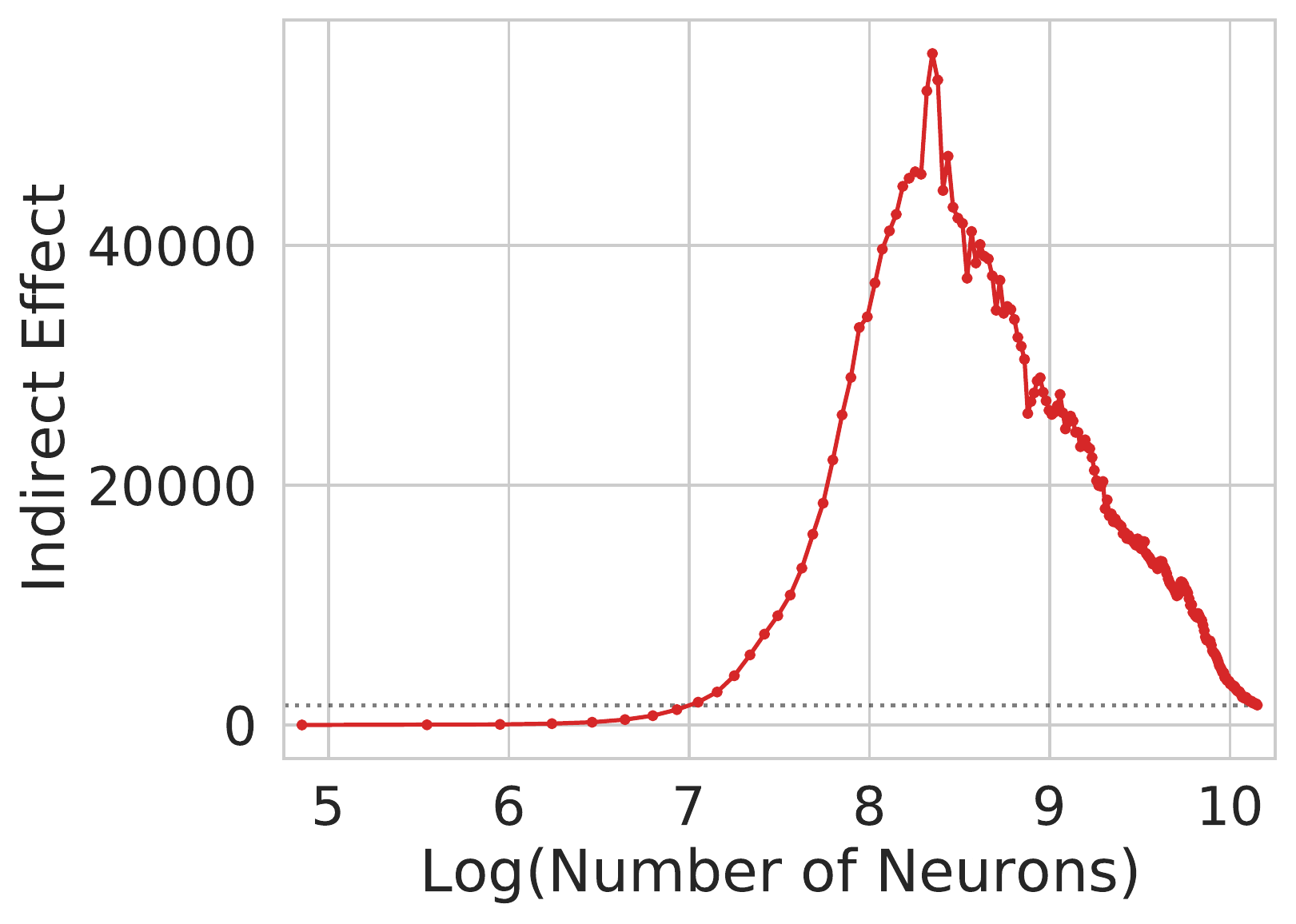}
        \caption{Simple agreement}
    \end{subfigure}
    \vspace{0.35cm}
    
    \begin{subfigure}{0.99\linewidth}
        \centering
        \includegraphics[width=0.63\linewidth]{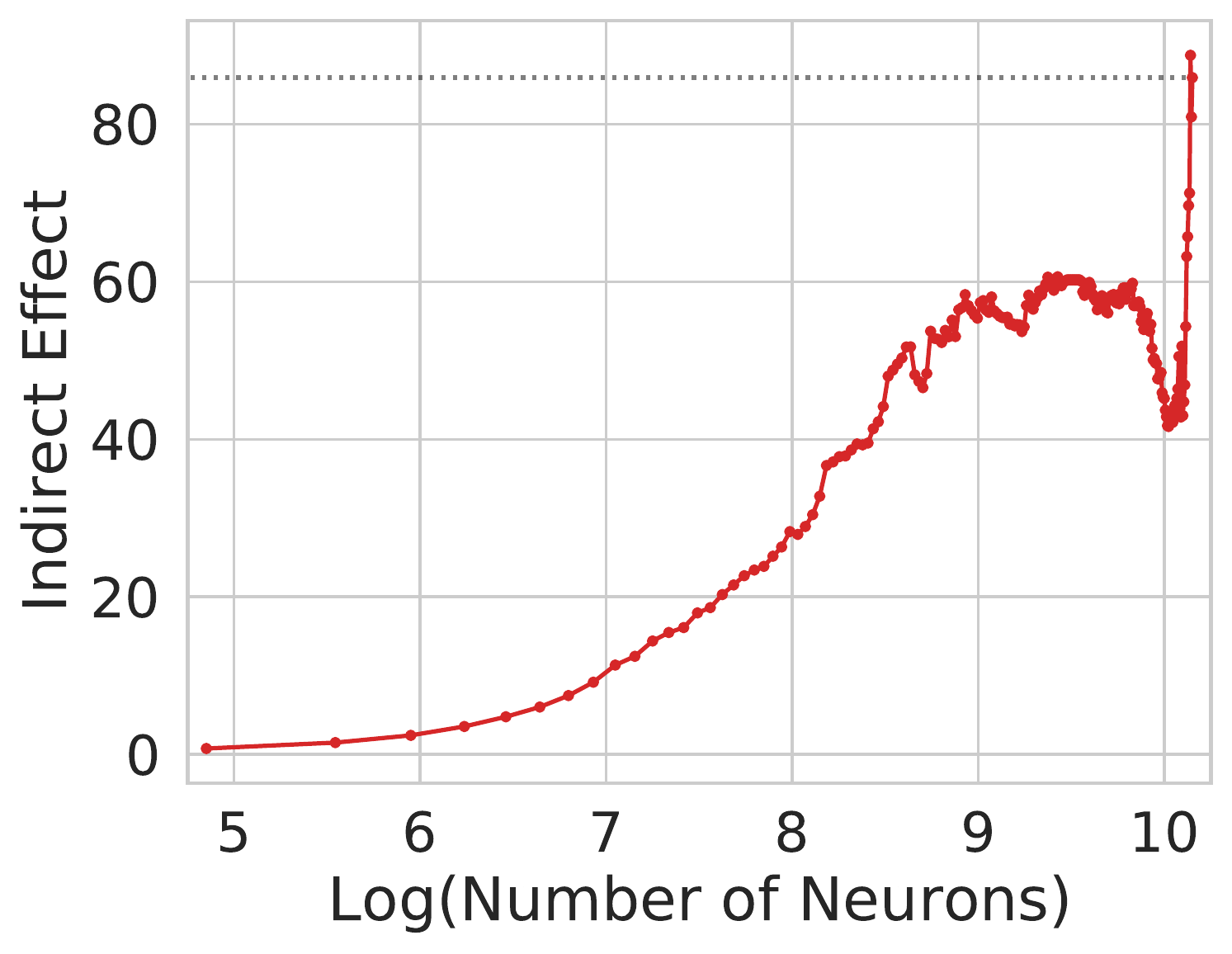}
        \caption{Across a singular RC}
    \end{subfigure}
    \caption{Indirect effects when intervening on increasing numbers of neurons in XGLM. The dashed line represents the total effect. For `simple agreement', there exists a set of neurons that strongly prefers grammatical completions; however, there are many more neurons that have weak preferences against them, and this results in the model as a whole having weak preferences for correct verb inflections. For `across a singular RC', however, almost every set of neurons seems to have preferences for grammatical inflections.}
    \label{fig:sparsity}
\end{figure}

We report two metrics: (1) the percentage of neurons at which we see the maximum NIE, and (2) the minimum percentage of neurons required for the NIE to reach the TE of the model. These correspond to the peak NIE and the point at which the NIEs cross the dashed line in Figure~\ref{fig:sparsity}.

For `simple agreement' (Table~\ref{tab:sparsity}), the proportion of neurons to reach the TE is generally small, especially for MLMs. However, the TE itself is often a couple orders of magnitude smaller for MLMs than for ALMs; thus, these percentages are not comparable across model architectures.

The proportion of neurons required to achieve the maximum NIE is typically lower for MLMs than ALMs. In other words, \textbf{syntax neurons are more sparse in masked language models than autoregressive language models.}\footnote{French is an exception: there are more syntax-sensitive neurons in both monolingual and multilingual models.}

The percentage of neurons to reach the maximum NIE does not significantly differ across monolingual and multilingual models, however. This means that \textbf{multilingual models do not consistently encode syntactic agreement in a more sparse way than monolingual models}. This and our neuron overlap results suggest that multilingual models encode syntactic information in a similar way to monolingual models (including the proportion of neurons sensitive to syntax), though most syntax-sensitive neurons tend to be language-specific rather than shared across languages.

Sparsity also differs across syntactic structures. For `simple agreement', NIEs peak at around 5--20\% of neurons. For `across a singular RC', the addition of every $k$ neurons almost always increases the NIEs. \textbf{Long-range syntactic information seems to be distributed throughout the majority of neurons in XGLM, but short-range syntactic information is more sparsely encoded}.

These numbers hide more interesting trends, however. The TEs for mBERT are often close to 0 across structures, while the maximum NIEs are in the hundreds for those same structures.\footnote{The effect contours for mBERT have a similar contour to those in Figure~\ref{fig:sparsity}, though the TE ($\approx$0) and maximum NIE ($\approx$340) for `simple agreement' are far smaller.} This has interesting implications for interpreting behavioral analyses: studies such as \citet{hu2020systematic} and \citet{mueller2020cross} suggest that mBERT does not have strong syntax-sensitive preferences compared to autoregressive language models, and the low TEs we observe support this. However, this obscures that \textbf{there are actually many neurons in mBERT which are highly sensitive to syntactic agreement}, as indicated by the high maximum NIEs: we observe weak agreement preferences in the model as a whole because there are many more neurons which have weak preferences \emph{against} syntactic agreement (i.e., small negative NIEs), perhaps because those neurons are specializing in other phenomena (e.g., token collocations or semantic agreement). Thus, behavioral analyses of model behavior may be underestimating the sensitivity of models to syntactic phenomena, for there is negative interference from neurons that prefer non-syntax-sensitive completions.

\section{Discussion}
We observe two distinct layer-wise NIE patterns for syntactic agreement, depending on whether the subject and verb are adjacent or separated by other tokens. This extends the findings of \citet{finlayson2021causal} to multilingual MLMs and ALMs, as well as monolingual MLMs in various languages. Going beyond their findings, we ruled out the possibility that these neurons do not simply track semantic plausibility or bigram collocations more generally. While this is not conclusive evidence that these neurons are specialized to syntax, evidence from other behavioral and probing studies also supports the existence of neurons focused on syntax \citep{hewitt2019structural,elazar2021amnesic,bertgoldberg19}. \citet{cao2021diffmask} found neurons focused \emph{purely} on syntax, while \citet{tucker2022probe} found redundantly encoded syntactic information across neurons. It is not clear how much of the neuron overlap we observe is due to redundantly encoded information, but future work could investigate this.

A consistent trend across our experiments is that ALMs encode syntactic agreement in a distinct way from MLMs. In ALMs, there is more cross-lingual and cross-structure neuron overlap than in MLMs; more similar layer-wise effect patterns across structures and languages (though they are still distinct); and a greater proportion of neurons which are sensitive to agreement. This could be partially explained by ALMs' left-to-right processing of natural language input, which more closely resembles incremental inputs to human learners. MLMs are able to perform syntactic agreement \citep{hu2020systematic,bertgoldberg19}, but their fill-in-the-blank pre-training objectives may induce distinct representations of sentence structure as compared to models that process or predict inputs incrementally.

Why do we observe different indirect effect contours for short- and long-range agreement? Perhaps syntactic agreement is encoded using a single mechanism, but the way that syntactic information is used for predicting output tokens depends on the structure of the input or prior output tokens. Alternatively, there could be two completely distinct agreement mechanisms that function in different ways entirely. While our findings do not disambiguate between these possibilities (or some other separate type or amount of mechanisms), future work could employ methods like those in \citet{meng2022rome} to observe this distinction more explicitly. The findings of \citet{meng2022rome} suggest that the model regions that are implicated in \emph{saying} something are distinct from those implicated in \emph{knowing} something---that is, knowledge retrieval and predicting particular tokens are separate mechanisms in pre-trained language models. Perhaps their method could be extended to study syntactic agreement, such that we can better understand what, exactly, these distinct indirect effect trends represent.

\section{Conclusions}
We have used causal mediation analysis to observe which neurons track syntactic agreement in multilingual pre-trained language models, and in which layers they are concentrated. We found two distinct layer-wise contours for syntactic agreement regardless of the language, multilinguality, or architecture of the model (\S\ref{sec:results_nie});
that syntax-sensitive neurons are shared across languages in autoregressive language models (\S\ref{sec:overlap_lang});
and that the neuron sparsity of syntactic agreement is similar in monolingual and multilingual models (\S\ref{sec:sparsity}).
We also found that behavioral analyses of masked language models obscure the extent to which their neurons are sensitive to syntactic agreement (\S\ref{sec:sparsity}).

\section*{Acknowledgments}
We thank the members of NYU's Computation and Psycholinguistics Lab for valuable feedback on earlier versions of this work.

This material is based upon work supported by the National Science Foundation (NSF) under Grant No. BCS-2114505. Aaron Mueller was supported by a National Science Foundation Graduate Research Fellowship (Grant \#1746891). This work was also supported by supported in part through the NYU IT High Performance Computing resources, services, and staff expertise.

\bibliography{anthology,custom}
\bibliographystyle{acl_natbib}

\appendix

\section{Example Sentences}\label{sec:examples}
Here, we present examples of each syntactic structure we observe in each language.

\ex.\ \textit{Simple agreement (English)}:
    \a. The woman observes/*observe.

\ex.\ \textit{Simple agreement (French)}:
    \ag. L' homme approuve/*approuvent. \\
        The man approves/*approve.\\

\ex.\ \textit{Simple agreement (German)}:
    \ag. Der Arzt weiß/*wissen.\\
        The physician knows/*know.\\

\ex.\ \textit{Simple agreement (Dutch)}:
    \ag. De schrijver begrijpt/*begrijpen.\\
        The writer understands/*understand.\\

\ex.\ \textit{Simple agreement (Finnish)}:
    \ag. Täti ymmärtää/*ymmärtävät.\\
        Aunt understands/*understand.\\
        ``The aunt understands/*understand.''

\begin{figure*}
    \centering
    \includegraphics[width=0.75\linewidth]{indirect_effects/legend.png}
    \vspace{-.25em}
    
    \includegraphics[width=0.275\linewidth,trim={0 0 0 0.1cm},clip]{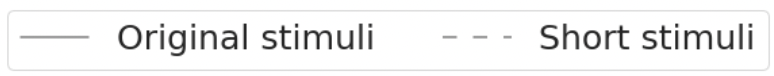}
    
    \begin{subfigure}{0.19\linewidth}
        \includegraphics[height=2.5cm,trim={0 0 13.4cm 0},clip]{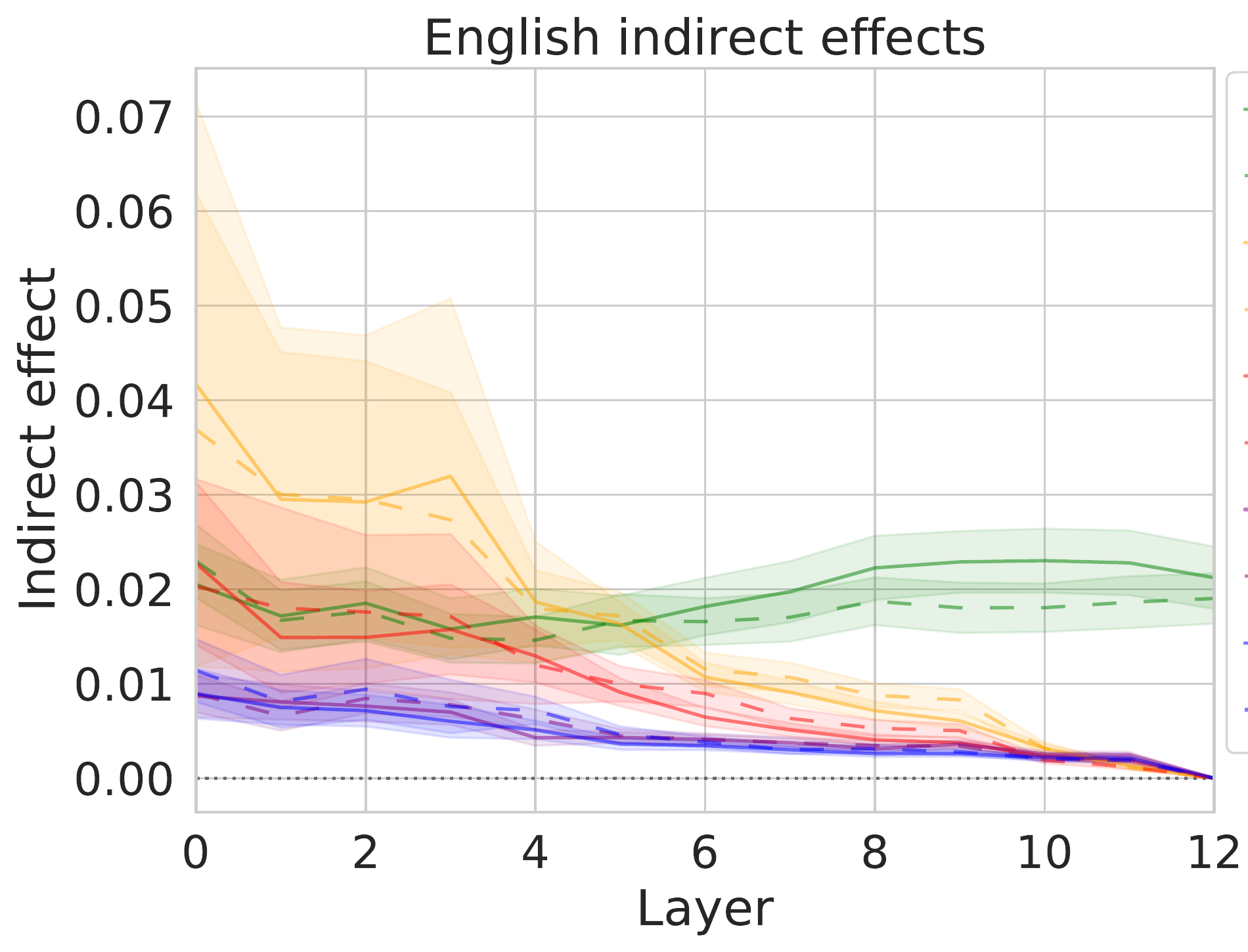}
        \caption{BERT}
    \end{subfigure}
    \hfill
    \begin{subfigure}{0.19\linewidth}
        \includegraphics[height=2.5cm,trim={1.25cm 0 16.6cm 0},clip]{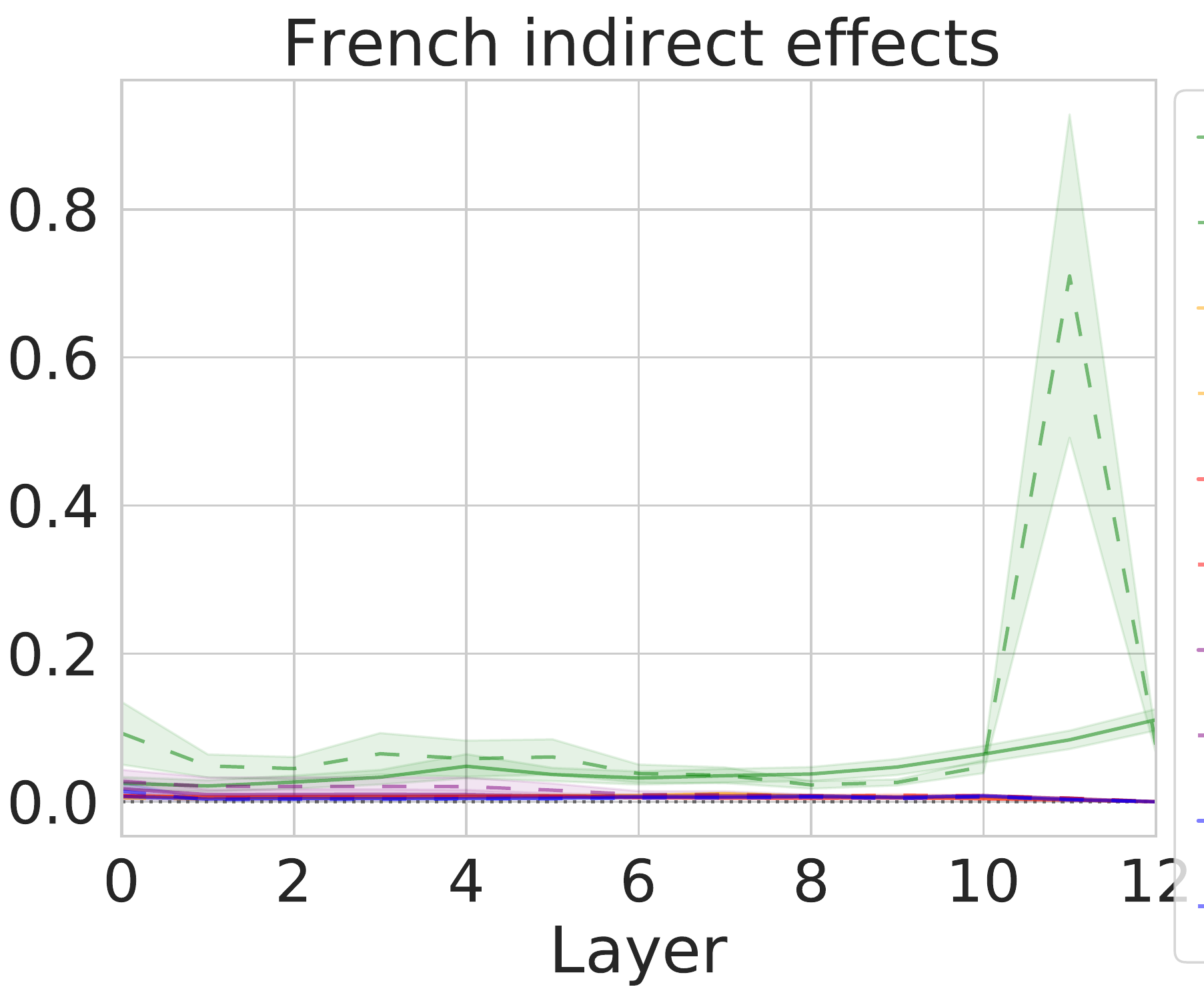}
        \caption{CamemBERT}
    \end{subfigure}
    \hfill
    \begin{subfigure}{0.19\linewidth}
        \includegraphics[height=2.5cm,trim={1.25cm 0 14.5cm 0},clip]{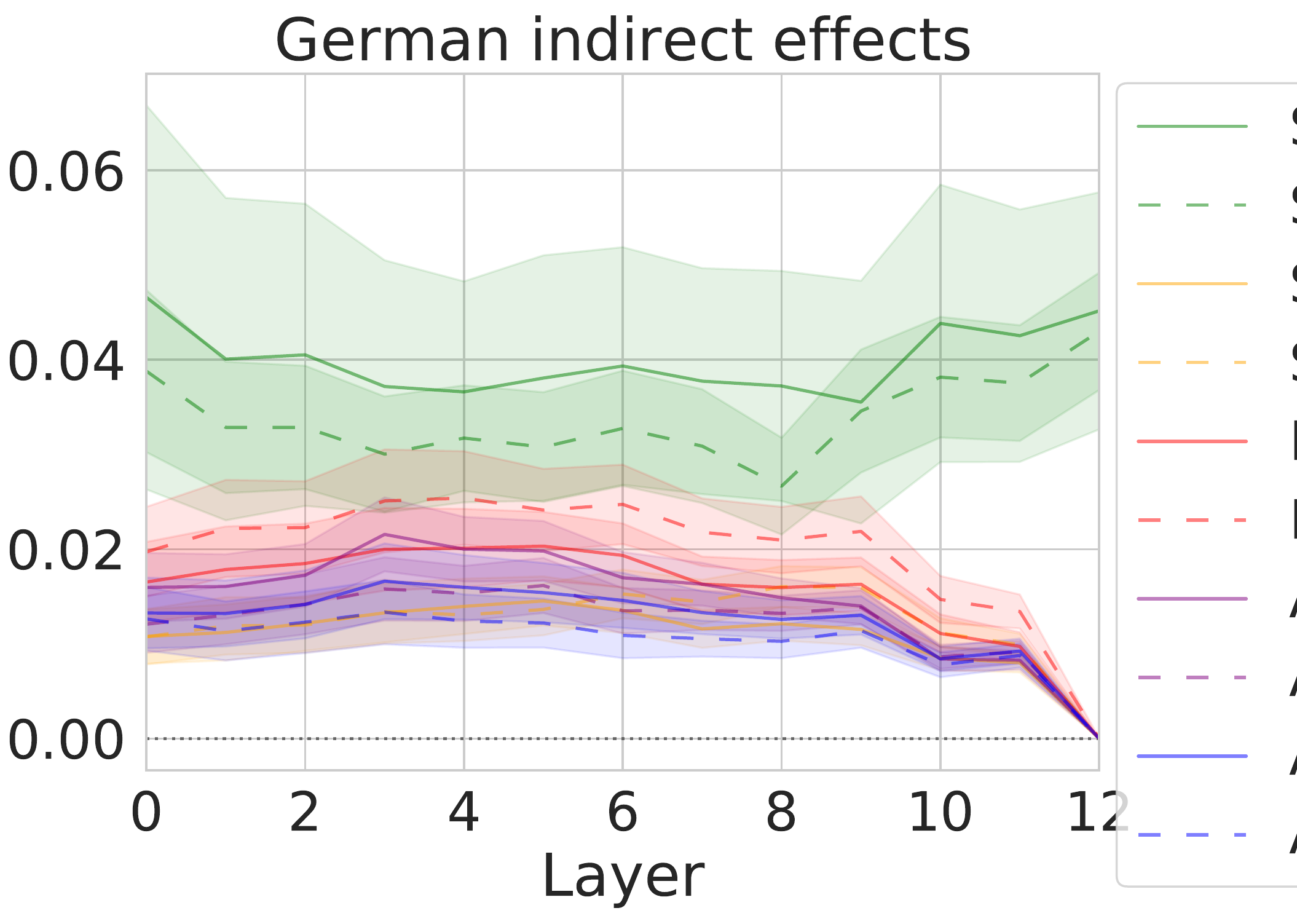}
        \caption{GermanBERT}
    \end{subfigure}
    \begin{subfigure}{0.19\linewidth}
        \includegraphics[height=2.5cm,trim={1.25cm 0 14.2cm 0},clip]{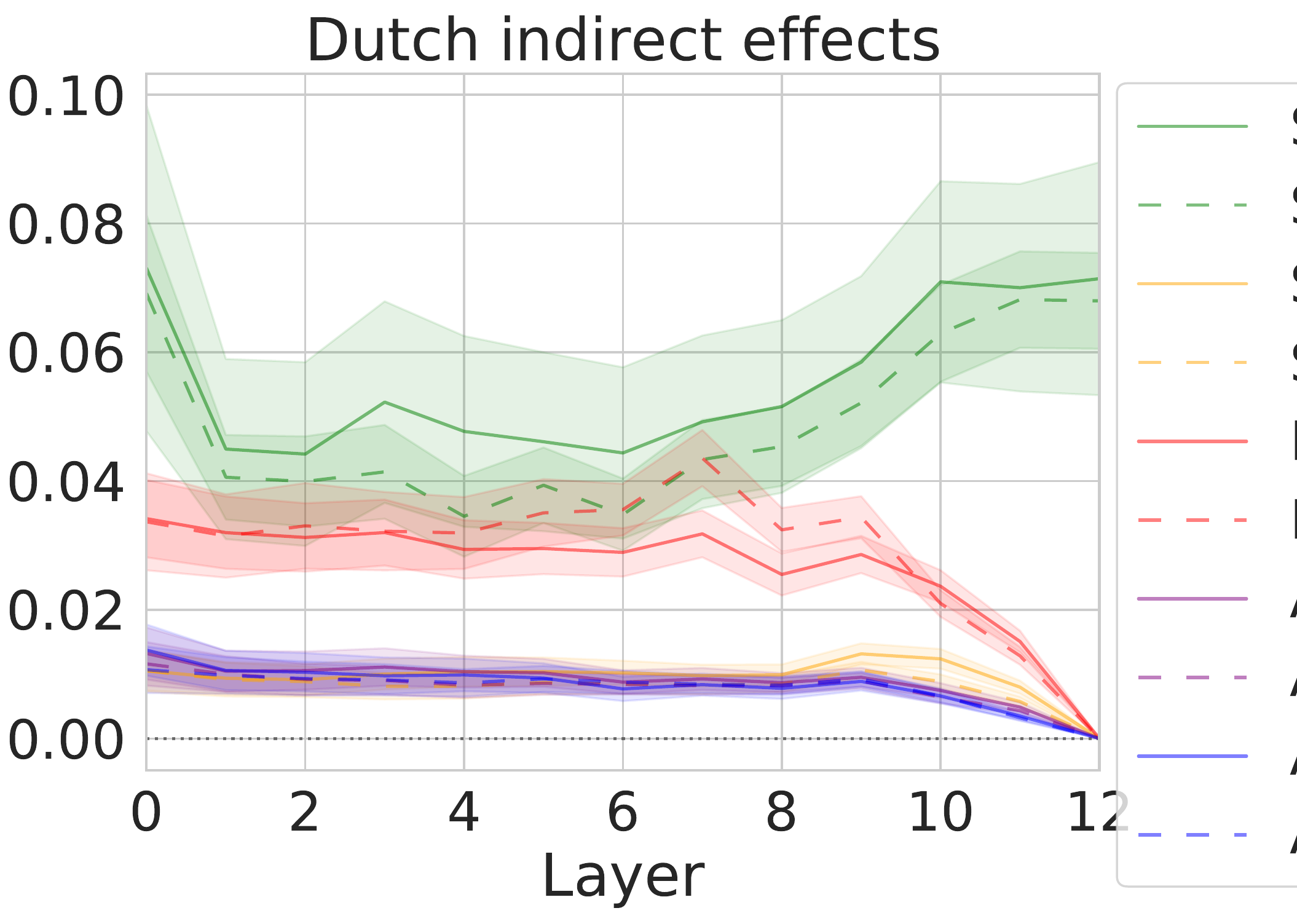}
        \caption{BERTje}
    \end{subfigure}
    \begin{subfigure}{0.19\linewidth}
        \includegraphics[height=2.5cm,trim={1.25cm 0 16.4cm 0},clip]{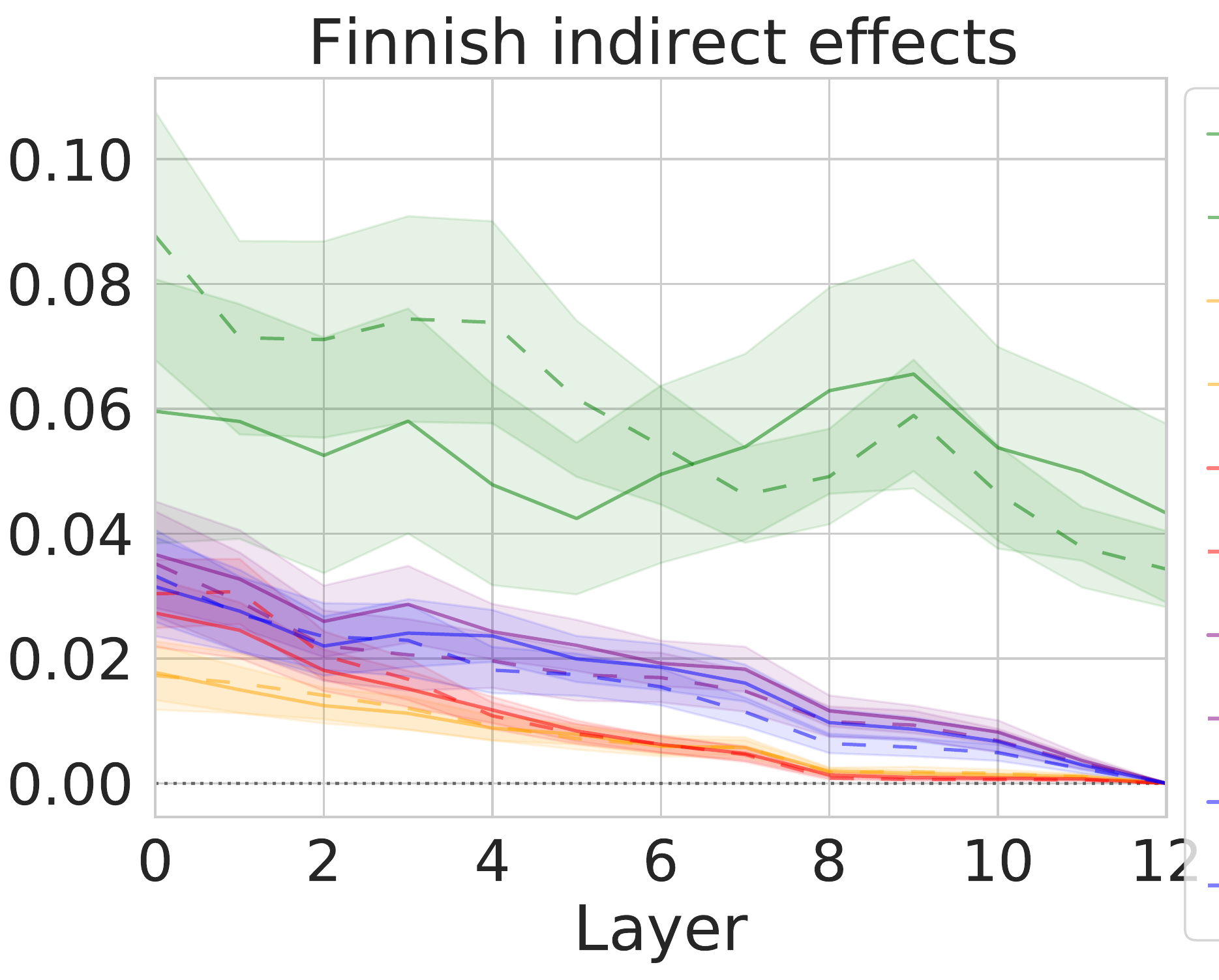}
        \caption{FinnishBERT}
    \end{subfigure}
    \caption{Natural indirect effects for the top 5\% of neurons in each layer for monolingual masked language models. The indirect effect contours we observe do not vary significantly when replacing the nouns and verbs with shorter, more frequent words---except in layer 11 of CamemBERT.}
    \label{fig:long_vs_short_stimuli}
\end{figure*}

For each of the following syntactic structures containing a grammatical number attractor, we separate structures by whether the attractor is singular or plural. For concision, we simply present examples of each structure without separating out examples by the number of the attractor. Note that Finnish mainly uses \emph{postpositions} rather than prepositions; the attractor still intervenes between the main subject and its verb, but the order of the preposition and noun phrase is different compared to the Indo-European languages we consider.

\ex.\ \textit{Across a relative clause (English)}:
    \a. The woman that the guards like observes/*observe.

\ex.\ \textit{Across a relative clause (French)}:
    \ag. L' homme que le chef suit approuve/*approuvent. \\
        The man that the boss follows approves/*approve.\\

\ex.\ \textit{Across a relative clause (German)}:
    \ag. Der Arzt den die Tiere vergeben weiß/*wissen.\\
        The physician that the animals forgive knows/*know.\\

\ex.\ \textit{Across a relative clause (Dutch)}:
    \ag. De schrijver die de ouder roept begrijpt/*begrijpen.\\
        The writer that the parent calls understands/*understand.\\

\ex.\ \textit{Across a relative clause (Finnish)}:
    \ag. Täti jota luistelijat kehuvat ymmärtää/*ymmärtävät.\\
        Aunt that skaters praise understands/*understand.\\
        ``The aunt that the skaters praise understands/*understand.''
        
\ex.\ \textit{Across a prepositional phrase (English)}:
    \a. The woman behind the cars observes/*observe.

\ex.\ \textit{Across a prepositional phrase (French)}:
    \ag. L' homme devant le chat approuve/*approuvent. \\
        The man in-front-of the cat follows approves/*approve.\\

\ex.\ \textit{Across a prepositional phrase (German)}:
    \ag. Der Arzt nahe den Äpfeln weiß/*wissen.\\
        The physician near the apples knows/*know.\\

\ex.\ \textit{Across a prepositional phrase (Dutch)}:
    \ag. De schrijver achter de fiets begrijpt/*begrijpen.\\
        The writer behind the bike understands/*understand.\\

\ex.\ \textit{Across a \textbf{post}positional phrase (Finnish)}:
    \ag. Täti puiden lähellä ymmärtää/*ymmärtävät.\\
        Aunt trees near understands/*understand.\\
        ``The aunt near the trees understands/*understand.''

\begin{figure}[t]
    \centering
    \begin{subfigure}{0.9\linewidth}
        \includegraphics[width=0.9\linewidth]{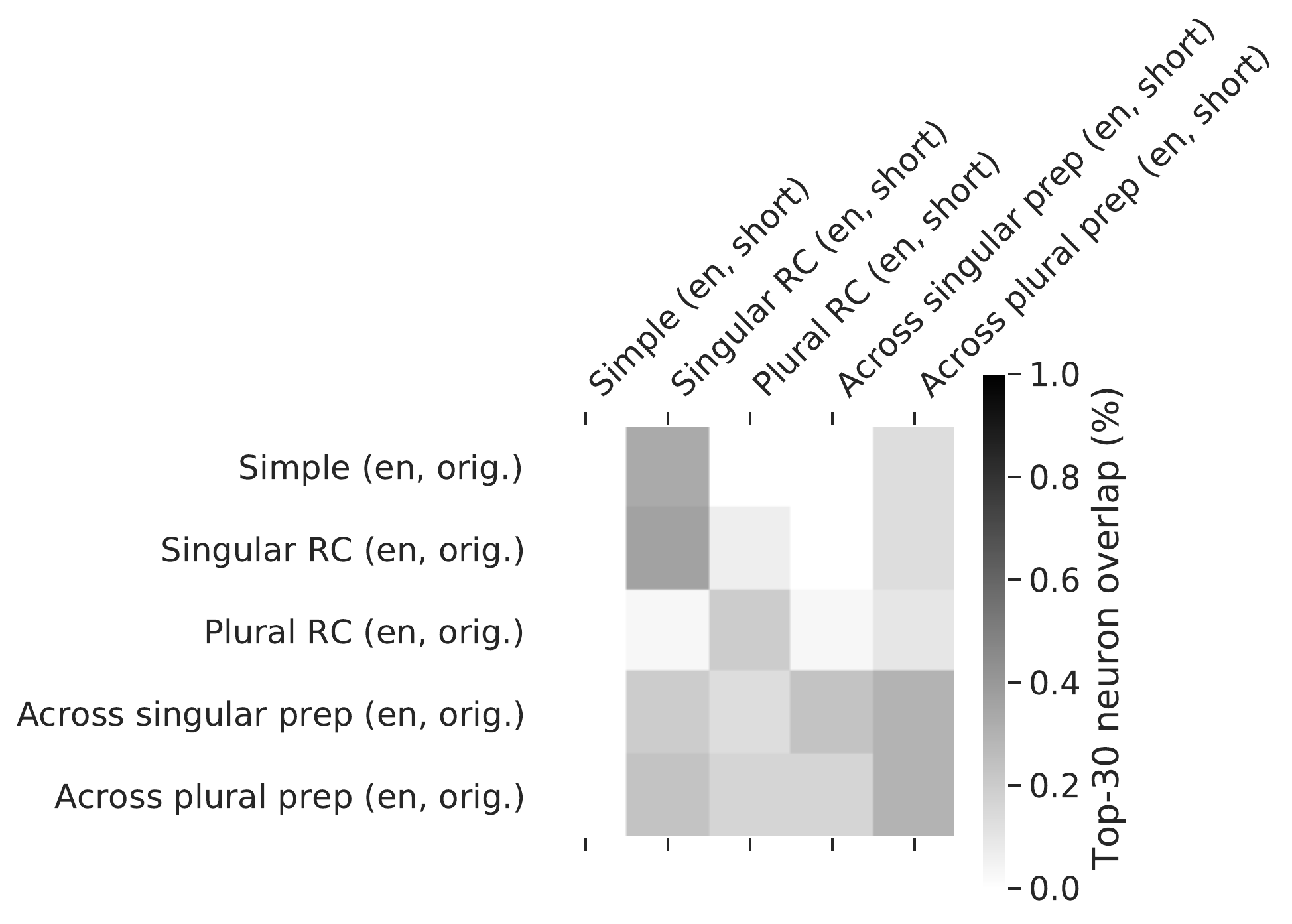}
        \caption{mBERT}
    \end{subfigure}
    \begin{subfigure}{0.9\linewidth}
        \includegraphics[width=0.9\linewidth]{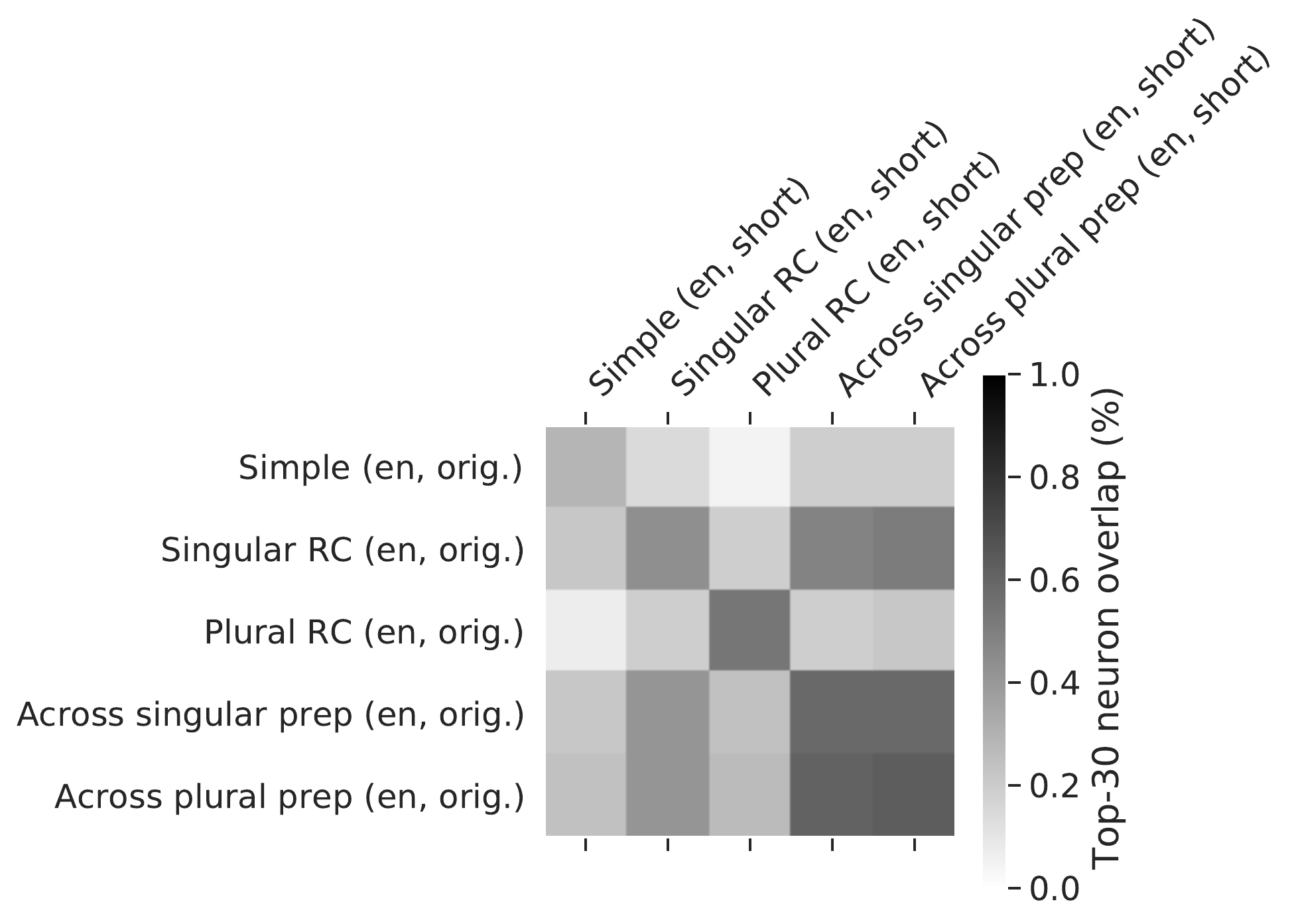}
        \caption{XGLM}
    \end{subfigure}
    \caption{Neuron overlap across structures (including baselines) in English for (a) mBERT and (b) XGLM. There is significant overlap between the original stimuli and short-word stimuli, though this is more the case for XGLM than mBERT.}
    \label{fig:overlap_short_vs_orig}
\end{figure}

\section{Invariance to Short- and Long-Word Stimuli}\label{app:short_long_stimuli}
When using the stimuli from \citet{finlayson2021causal}, most of the subjects and verbs are split into multiple tokens. These are generally long and relatively infrequent nouns and verbs like ``managers'' and ``observe''. We could use more stimuli if we replace each word with words that are shorter and more frequent in pre-training corpora, such as ``cats'' and ``see''.

Will these lexical replacements change the trends we observe? We observe the layer-wise natural indirect effect of the top neurons in each layer for the original stimuli and the short-word stimuli to see if lexical replacements have an effect on the way neurons encode syntactic agreement in monolingual BERT models. Our results (Figure~\ref{fig:long_vs_short_stimuli}) are nearly identical for the original stimuli and the short stimuli. A notable exception is layer 11 of CamemBERT, where indirect effects are so large that the rest of the effects are dwarfed by comparison. However, when excluding this result, indirect effect contours look similar between original and short stimuli.

\begin{figure*}
    \centering
    
    (a) mBERT:
    \begin{subfigure}{0.28\linewidth}
        \includegraphics[width=\linewidth]{neuron_overlap/neuron_overlap_en-fr_bert-base-multilingual-cased.pdf}
    \end{subfigure}
    \hfill
    \begin{subfigure}{0.28\linewidth}
        \includegraphics[width=\linewidth]{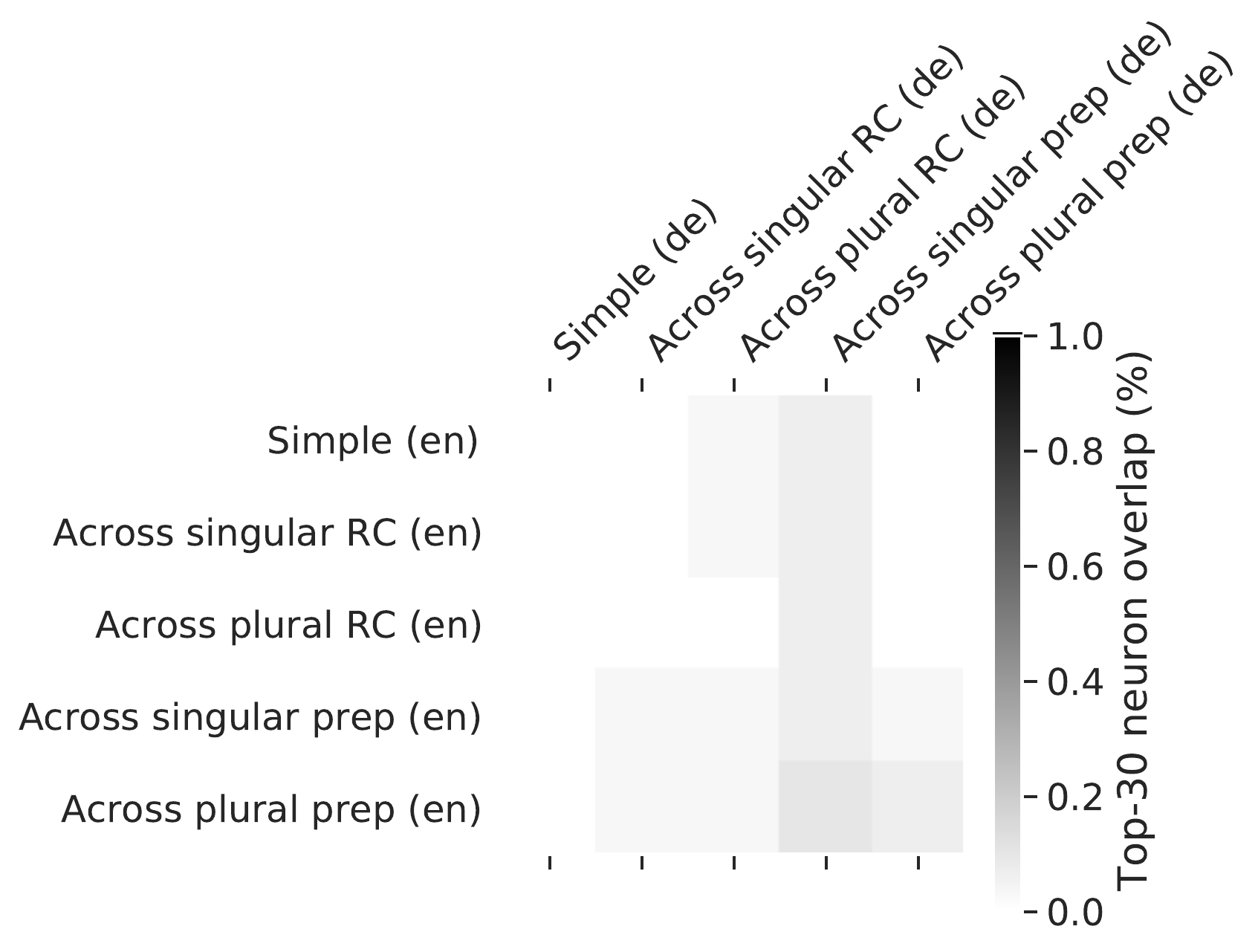}
    \end{subfigure}
    \hfill
    \begin{subfigure}{0.28\linewidth}
        \includegraphics[width=\linewidth]{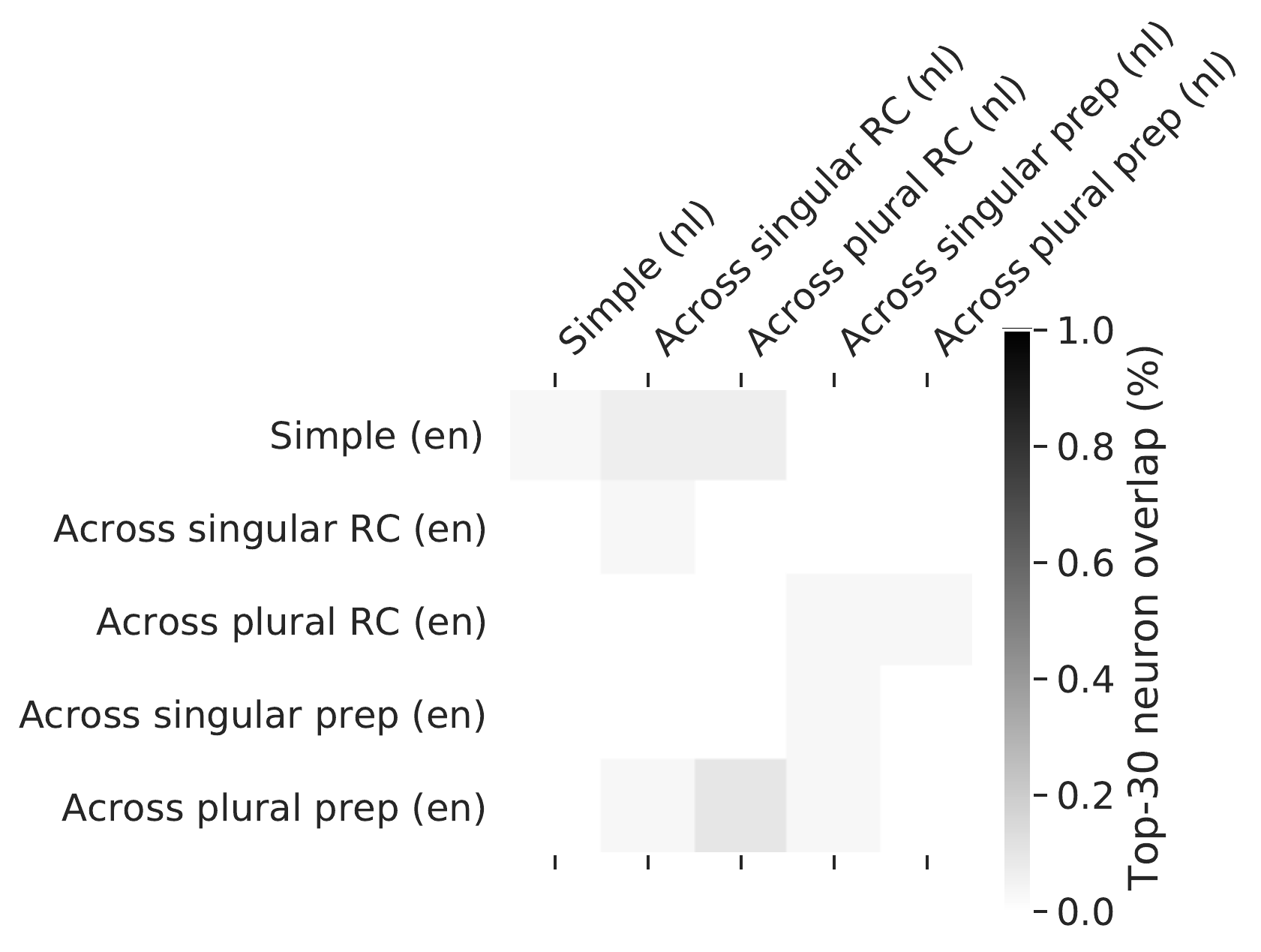}
    \end{subfigure}
    
    (b) XGLM:
    \begin{subfigure}{0.28\linewidth}
        \includegraphics[width=\linewidth]{neuron_overlap/neuron_overlap_en-fr_xglm-564M.pdf}
    \end{subfigure}
    \begin{subfigure}{0.28\linewidth}
        \includegraphics[width=\linewidth]{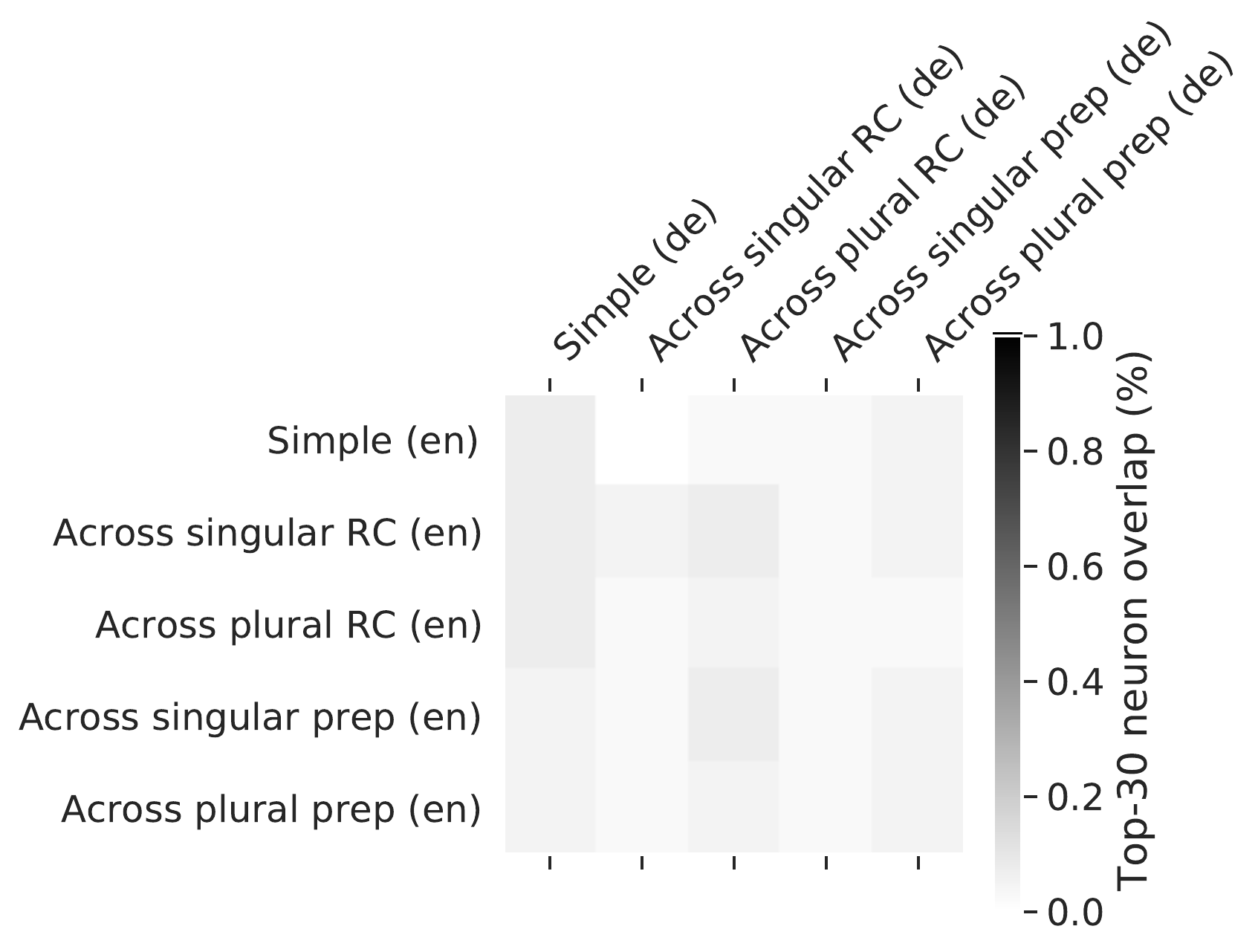}
    \end{subfigure}
    \begin{subfigure}{0.28\linewidth}
        \includegraphics[width=\linewidth]{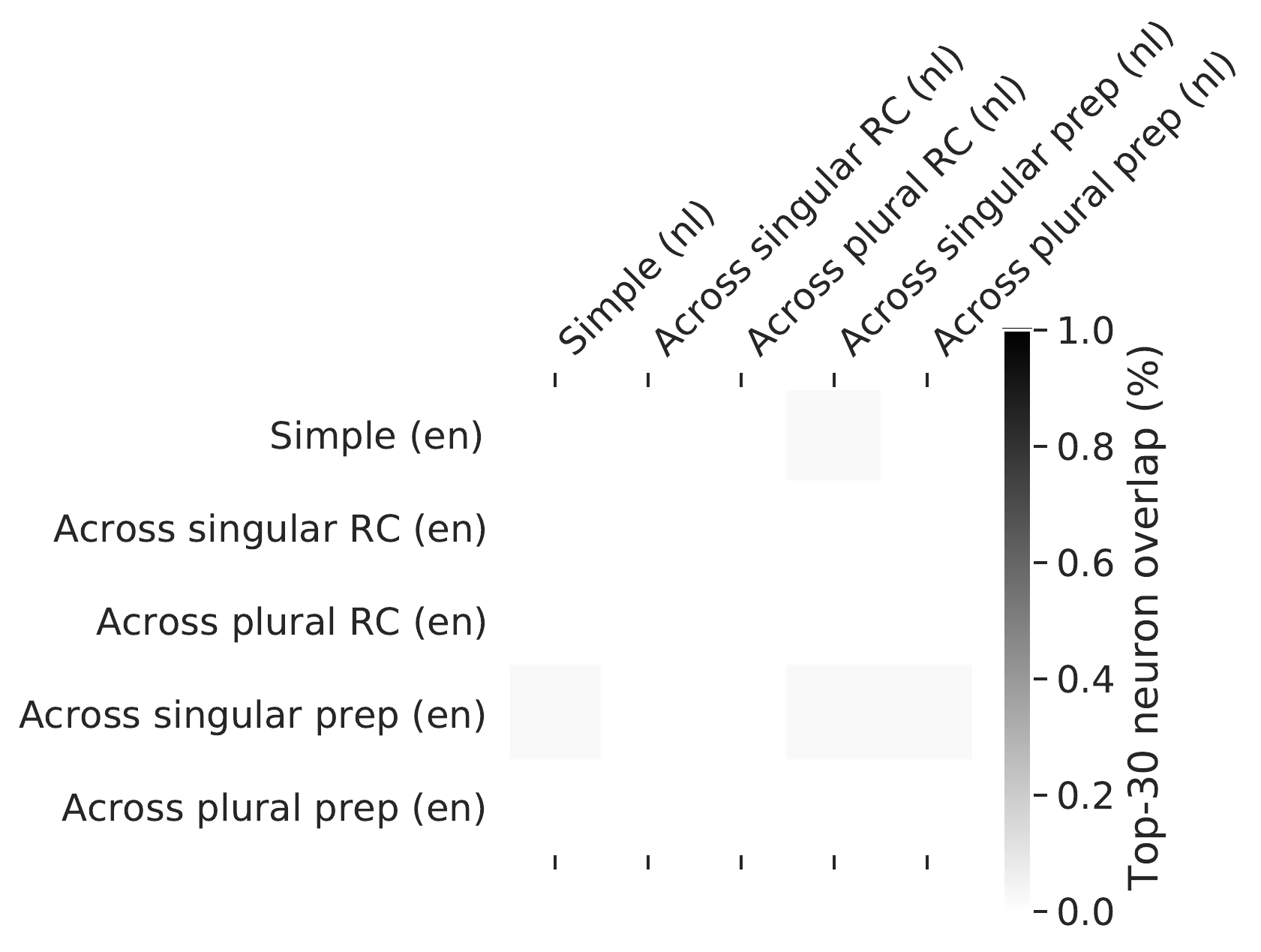}
    \end{subfigure}
    \caption{Neuron overlap for the top 30 neurons in mBERT (top row) and XGLM (bottom row). We show overlaps between English and French (left), German (center), and Dutch (right).}
    \label{fig:neuron_overlap_crossling}
\end{figure*}

We also compare the extent of neuron overlap between original and short stimuli for multilingual BERT and XGLM. Our results (Figure~\ref{fig:overlap_short_vs_orig}) show a relatively high degree of overlap, especially for XGLM. However, overlap is somewhat lower than when we use only one stimulus type (Figure~\ref{fig:overlap_en}). Ideally, overlap should be nearly 100\% along the diagonal of both matrices if these neurons account only for syntactic agreement rather than specific lexical items, so these results suggest that lexical (and not syntactic) features may account for a notable proportion of the neuron overlap we observe in our previous experiments. Alternatively, it could mean that these neurons attend both syntactic \emph{and} lexical information. Nonetheless, overlaps are still significant and indirect effects still look similar when swapping our nouns and verbs, so it is likely that models are picking up on some abstraction for syntactic agreement that generalizes across specific token sequences.

These results suggest that the neuron-level effects we observe are not simply spurious lexical correlations. More significantly, this is further evidence that \textbf{the neuron-level effects we observe are not word-level effects, but some more abstract structural feature(s) that the model has learned}.

\section{Neuron Overlap Across Languages: Full Results}\label{app:neuron_overlap_all}
Here, we present neuron overlaps across languages for mBERT and XGLM (Figure~\ref{fig:neuron_overlap_crossling}). As in \S\ref{sec:overlap_lang}, we present overlaps for the top 30 neurons (in \emph{any} layer of the model) per structure per language. As before, we find that neuron overlap is generally greater in autoregressive LMs than masked LMs.

Neuron overlaps are most prominent between English and French; while not typologically the most closely related language pair, English and French share a great deal of vocabulary and have similar SVO word orders when pronominal objects are not present. German, meanwhile, uses SOV with V2 in main clauses.

\section{Limitations}
Perhaps the greatest limitation of our method---and many other causal probing methods \citep{vig2020causal,finlayson2021causal,ravfogel2021counterfactual}---is that we are limited to stimuli where the subjects on which we intervene and the competing verb forms are one token each. This greatly limits the range of subjects and verbs (and languages) we can consider in this study, especially for more multilingual models where a greater proportion of words are split into subwords by the tokenizer. Models may use a different mechanism altogether to calculate the probability of two competing verbs given the presence or lack of a morpheme like \{-s\} which expresses number information, and our method would not allow us to understand where and how models are performing this kind of agreement. While one can, in theory, compare the probability of variable-length token sequences in autoregressive language models, there is no principled way to do this in masked language models. And in practice, autoregressive language models tend to prefer shorter sequences. Future work could consider probing methods which allow for variable-length span predictions.

There are also more general issues with probing individual neurons. Complex phenomena like syntactic agreement are likely to be encoded in \emph{sets of} neurons, rather than individual neurons; indeed, we find evidence for this in \S\ref{sec:sparsity}. This means that analyzing individual neurons can result in oversimplified understandings of where and how certain phenomena are encoded and used. Future causal probing work could focus on non-parametric methods which allow one to probe multiple neurons simultaneously, such that we may causally implicate model \emph{regions} rather than just individual components like neurons or attention heads.

\end{document}